\def\year{2022}\relax
\documentclass[letterpaper]{article} 
\usepackage{aaai22}  
\usepackage{times}  
\usepackage{helvet}  
\usepackage{courier}  
\usepackage[hyphens]{url}  
\usepackage{graphicx} 
\usepackage{natbib}  
\usepackage{caption} 
\DeclareCaptionStyle{ruled}{labelfont=normalfont,labelsep=colon,strut=off} 
\usepackage{algorithm}
\usepackage[noend]{algorithmic}
\usepackage{bm}
\usepackage{multirow}
\usepackage{booktabs}
\usepackage{longtable}
\usepackage{hyperref}       
\usepackage{url}            

\usepackage{newfloat}
\usepackage{listings}
\floatstyle{ruled}
\newfloat{listing}{tb}{lst}{}
\floatname{listing}{Listing}
\title{Detecting Misclassification Errors in Neural Networks\\ with a Gaussian Process Model}
\author {
	Xin Qiu,\textsuperscript{\rm 1}
	Risto Miikkulainen \textsuperscript{\rm 1, 2}
}
\affiliations {
	\textsuperscript{\rm 1} Cognizant AI Labs\\
	\textsuperscript{\rm 2} The University of Texas at Austin\\
	qiuxin.nju@gmail.com, risto@cognizant.com
}

\begin{document}

\maketitle

\begin{abstract}
As neural network classifiers are deployed in real-world applications, it is crucial that their failures can be detected reliably. One practical solution is to assign confidence scores to each prediction, then use these scores to filter out possible misclassifications. However, existing confidence metrics are not yet sufficiently reliable for this role. This paper presents a new framework that produces a quantitative metric for detecting misclassification errors. This framework, RED, builds an error detector on top of the base classifier and estimates uncertainty of the detection scores using Gaussian Processes. Experimental comparisons with other error detection methods on 125 UCI datasets demonstrate that this approach is effective. Further implementations on two probabilistic base classifiers and two large deep learning architecture in vision tasks further confirm that the method is robust and scalable. Third, an empirical analysis of RED with out-of-distribution and adversarial samples shows that the method can be used not only to detect errors but also to understand where they come from. RED can thereby be used to improve trustworthiness of neural network classifiers more broadly in the future.
\end{abstract}

\section{Introduction}
\label{sec:introduction}
Classifiers based on Neural Networks (NNs) are widely deployed in many real-world applications \citep{LeCun2015, Anjos2015, Alghoul2018, Shahid2019}. Although good prediction accuracies are achieved, it is usually not clear whether a particular prediction can be trusted, which is a severe issue especially in safety-critical domains such as healthcare \citep{Dan2018, Gupta2007, Shahid2019}, finance \citep{Dixon2017}, and automated driving \citep{Janai2017, Hecker2018}.

One way to estimate trustworthiness of a classifier prediction is to use its inherent confidence-related score, e.g., the maximum class probability \citep{Hendrycks2017}, entropy of the softmax outputs \citep{Williams97}, or difference between the highest and second highest activation outputs \citep{Monteith2010}. However, these scores are unreliable and may even be misleading: Predictions often have high-confidence but are nevertheless incorrect \citep{Provost98, Guo17, Nguyen15, Goodfellow14, Amodei16}. In a practical setting, it is beneficial to have a detector that can raise a red flag whenever the predictions are likely to be wrong. A human observer can then evaluate such predictions, making the classification system safer.

Such a detector can be constructed by first developing quantitative metrics for measuring the reliability of predictions under different circumstances, and then setting a warning threshold based on users' preferred precision-recall tradeoff.  Existing such methods can be categorized into three types based on their focus: error detection, which aims to detect the natural misclassifications made by the classifier \citep{Hendrycks2017, Jiang18, Charles19}; out-of-distribution (OOD) detection, which identifies samples that are from different distributions compared to training data \citep{liang2018, lee2018, Devries2018}; and adversarial sample detection, which filters out samples from adversarial attacks \citep{Lee18, Wang19, Jonathan19}.

Among these categories, error detection, also called misclassification detection \citep{Jiang18} or failure prediction \citep{Charles19}, is the most challenging \citep{Jonathan19} and most underexplored. For instance, \citet{Hendrycks2017} defined a baseline error detection score using the maximum class probability after the softmax layer. Although this baseline is a good starting point, it is ineffective in some cases, indicating that there is room for improvement \citep{Hendrycks2017}. \citet{Jiang18} proposed Trust Score, which measures the similarity between the original classifier and a modified nearest-neighbor classifier. The main limitation of this method is scalability: the Trust Score may provide no or negative improvement over the baseline for high-dimensional data. ConfidNet \citep{Charles19} builds a separate NN model to learn the true class probablity, i.e.\ softmax probability for the ground-truth class. However, ConfidNet itself is a standard NN, so its detection scores may be unreliable or misleading: A random input may generate a random detection score, and ConfidNet does not indicate uncertainty of these detection scores. Moreover, none of these methods can differentiate natural classifier errors from risks caused by OOD or adversarial samples; if a detector could do that, it would be easier for practitioners to fix the problem, e.g., by retraining the original classifier or by applying better preprocessing techniques to filter out OOD or adversarial data.

To meet these challenges, a new framework is developed in this paper for error detection in NN classifiers. The main idea is to learn to predict the correctness of a classification result with a Gaussian Process (GP) model. The new system, referred to as RED (Residual-based Error Detection), not only produces an enhanced error detection score based on the original maximum class probability, but also provides a quantitative uncertainty estimation of that score. As a result, misclassification errors can be detected more reliably. Note that in this manner, RED is different from traditional confidence calibration methods \citep{Platt99, Zadrozny01, Zadrozny02, Guo17}, which do not improve misclassification detection.

The GP model in RED is constructed based on the RIO method for uncertainty estimation \citep[Residual prediction with Input/Output kernel;][]{Qiu2020}. It is notable that the modified RIO is only a part of the RED framework, and RED has several fundamental differences compared to RIO. First, whereas the original RIO is only applicable to regression models, RED works with classification tasks, which are more common. Second, whereas RIO calibrates the outputs of base models (thus reducing prediction errors), RED generates a new detection score for error detection usage only; the original classifier outputs are unchanged, and the classification accuracy remains the same. Third, whereas RIO quantifies the predictive uncertainty of the base model, RED only quantifies the uncertainty of the detection score, which is separate from the base classifier.

RED is compared empirically to state-of-the-art error detection methods on 125 UCI datasets and two vision tasks, with implementations on one standard NN classifier, two probabilistic NN classifiers, and two deep NN architectures. The results demonstrate that the approach is effective and robust. A further empirical study with OOD and adversarial samples shows the potential of using RED to diagnose the sources of mistakes as well, thereby paving the way to a comprehensive approach for improving trustworthiness of neural network classifiers in the future.

\section{Related Work}
\label{sec:relatedwork}
In the past two decades, a large volume of work was devoted to calibrating the confidence scores returned by classifiers. Early works include Platt Scaling \citep{Platt99, Mizil05}, histogram binning \citep{Zadrozny01}, isotonic regression \citep{Zadrozny02}, with recent extensions like Temperature Scaling \citep{Guo17}, Dirichlet calibration \citep{Kull19}, and distance-based learning \citep{Xing2020}. These methods focus on reducing the difference between reported class probability and true accuracy, and generally the rankings of samples are preserved after calibration. As a result, the separability between correct and incorrect predictions is not improved. In contrast, RED aims at deriving a score that can differentiate incorrect predictions from correct ones better.

A related direction of work is the development of classifiers with rejection/abstention option. These approaches either introduce new training pipelines/loss functions \citep{Bartlett08, Yuan10, Cortes16}, or define mechanisms for learning rejection thresholds under certain risk levels \citep{Bernard93, Santos05, Chow06, Geifman17}. In contrast, RED assumes an existing pretrained NN classifier, and provides an additional metric for detecting potential errors made by this classifier, without specifying a rejection threshold.

Designing metrics for detecting potential risks in NN classifiers has become popular recently. While most approaches focus on detecting OOD \citep{liang2018, lee2018, Devries2018} or adversarial examples \citep{Lee18, Wang19, Jonathan19}, work on detecting natural errors, i.e., regular misclassifications not caused by external sources, is more limited. \citet{Ortega95} and \citet{Koppel96} conducted early work in predicting whether a classifier is going to make mistakes, and \citet{Seewald01} built a meta-grading classifier based on similar ideas. However, these early works did not consider NN classifiers. More recently, \citet{Hendrycks2017} and \citet{Haldimann2019} demonstrated that raw maximum class probability is an effective baseline in error detection, although its performance was reduced in some scenarios. 

More elaborate techniques for error detection have also been developed recently. \citet{Mandelbaum2017} proposed a detection score based on the data embedding derived from the penultimate layer of a NN. However, their approach requires modifying the training procedure in order to achieve effective embeddings. \citet{Jiang18} introduced Trust Score to measure the similarity between a base classifier and a modified nearest-neighbor classifier. Trust Score outperforms the maximum class probability baseline in many cases, but negative improvement over baseline can be observed in high-dimensional problems, implying poor scalability of local distance computations. ConfidNet \citep{Charles19} learns to predict the class probability of true class with another NN, while Introspection-Net \citep{Jonathan19} utilizes the logit activations of the original NN classifier to predict its correctness. Since both models themselves are standard NNs, the detection scores returned by them may be arbitrarily high without any uncertainty information. Moreover, existing approaches for error detection cannot differentiate natural misclassification error from OOD or adversarial samples, making it difficult to diagnose the sources of risks. In contrast, RED explicitly reports its uncertainty about the estimated detection score, providing more reliable error detection. The uncertainty information returned by RED may also be helpful in clarifying the cause of classifier mistakes, as will be demonstrated in this paper.

\section{Methodology}
\label{sec:methodology}
This section gives the general problem statement, reviews the RIO method for uncertainty estimation, and describes the technical details of RED.
\subsection{Problem Statement}
\label{subsec:problem_statement}
Consider a training dataset $\mathcal{D}=(\mathcal{X},\mathbf{y})=\{(\mathbf{x}_i,y_i)\}_{i=1}^N$, and a pretrained NN classifier that outputs a predicted label $\hat{y}_i$ and class probabilities for each class $\sigma_i = [\hat{p}_{i,1}, \hat{p}_{i,2}, \dots, \hat{p}_{i,K}]$ given $\mathbf{x}_i$, where $N$ is the total number of training points and $K$ is the total number of classes.
The problem is to develop a metric that can serve as a quantitative indicator for detecting natural misclassification errors made by the pretrained NN classifier.

\subsection{RIO}
\label{subsec:RIO}
The RIO method \citep{Qiu2020} was developed to quantify point-prediction uncertainty in regression models. More specifically, RIO fits a GP to predict the residuals, i.e.\ the differences between ground-truth and original model predictions. It utilizes an I/O kernel, i.e.\ a composite of an input kernel and an output kernel, thus taking into account both inputs and outputs of the original regression model. As a result, it measures the covariances between data points in both the original feature space and the original model output space. For each new data point, a trained RIO model takes the original input and output of the base regression model, and predicts a distribution of the residual, which can be added back to the original model prediction to obtain both a calibrated prediction and the corresponding predictive uncertainty.

SVGP \citep{Hensman2013,Hensman2015} was used in RIO as an approximate GP to significantly reduce the computational cost. Both empirical results and theoretical analysis showed that RIO is able to consistently improve the prediction accuracy of base models as well as provide reliable uncertainty estimations. It therefore forms a promising foundation for improving reliability of error detection metrics as well.
\subsection{RED}
\label{subsec:RED}
Classification models generate class probabilities as their outputs. Therefore, the maximum class probability is an inherent baseline metric for error detection \citep{Hendrycks2017,Haldimann2019}. RED derives a more reliable detection score from this baseline by incorporating a GP model based on RIO. Since detecting errors in classification models is beyond the scope of the original RIO, two modifications are made to adapt RIO to this new domain.

First, since RIO was designed for single-output regression problems, it contains an output kernel only for scalar outputs. In RED, this original output kernel is extended to multiple outputs, i.e.\ to vector outputs such as those of the final softmax layer of a NN classifer, representing estimated class probabilities for each class. This modification allows RIO to access more information from the classifier outputs. This new variant of RIO is hereinafter referred to as mRIO (``m'' for multi-output).

Second, RIO estimates the point-prediction uncertainty by predicting the residuals between ground-truths and model outputs, both of which are limited to one-dimensional continuous space. However, ground-truth labels in classification problems are in a categorical space.
Therefore, a different problem needs to be constructed: Instead of learning to reach the ground-truths directly, RED learns to predict whether the original prediction is correct or not. A target detection score is assigned to each training data point according to whether it is correctly classified by the base model. The residual between this target score and the original maximum class probability is calculated, and an mRIO model is trained to predict these residuals. Given a new data point, the trained mRIO model combined with the base classifier thus provides an aggregated score for detecting misclassification errors. Note that the outputs of base classifiers are not changed.
\begin{figure*}[t]
	\centering
	\includegraphics[width=0.86\linewidth]{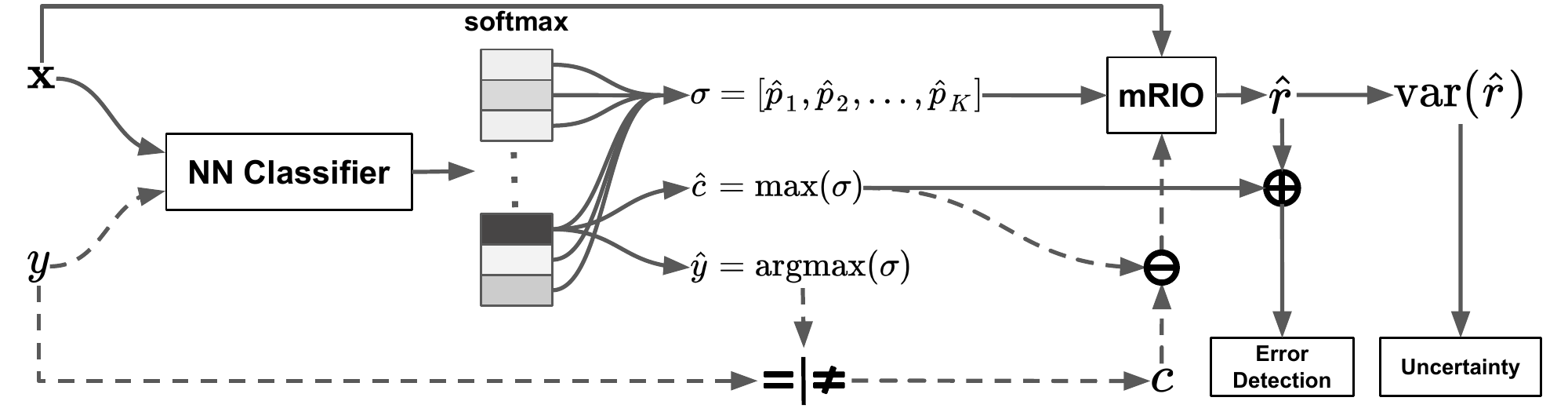}\\
	\caption{\textbf{The RED training and deployment processes.}  \label{fig:RED_flowchart}
		The solid pathways are active in both training and deployment phases, while the dashed pathways are active only in the training phase. During the training phase, a target detection score $c$ is assigned to each training sample according to whether it is correctly predicted by the original NN classifier or not. An mRIO model is then trained to predict the residual between the target detection score $c$ and the original maximum class probability $\hat{c}$. The I/O kernel in mRIO utilizes both the raw feature $\mathbf{x}$ and softmax outputs $\sigma$ to predict the residuals. In the deployment phase, given a new data point, the trained mRIO model provides a Gaussian distribution of estimated residual $\hat{r}$ defined by the mean $\bar{\hat{r}}$ and variance $\mathrm{var}(\hat{r})$. Addition of $\hat{r}$ and $\hat{c}$ forms a score for error detection, and $\mathrm{var}(\hat{r})$ indicates the corresponding uncertainty.
	}
\end{figure*}
\begin{algorithm}[t]
	\small
	\caption{RED training and deployment procedures}
	\label{alg:pseudo_RED}
	\begin{algorithmic}[1]
		\REQUIRE ${}$
		\\ $(\mathcal{X},\mathbf{y})=\{(\mathbf{x}_i,y_i)\}_{i=1}^N$: training data
		\\ $\mathbf{\hat{y}}=\{\hat{y}_i\}_{i=1}^N$: labels predicted by original NN classifier on training data
		\\ $\bm{\sigma}=\{\sigma_i = [\hat{p}_{i,1}, \hat{p}_{i,2}, \dots, \hat{p}_{i,K}]\}_{i=1}^N$: softmax outputs of original NN classifier on training data
		\\ $\mathbf{\hat{c}}=\{\hat{c}_i=\mathrm{max}(\sigma_i)\}_{i=1}^N$: maximum class probability returned by original NN classifier on training data
		\\ $\mathbf{x}_*$: data to be predicted
		\\ $\sigma_*$: softmax outputs of original NN classifier on $\mathbf{x}_*$
		\\ $\hat{c}_*$: maximum class probability returned by original NN classifier on $\mathbf{x}_*$
		\ENSURE ${}$
		\\ $\hat{c}^{\prime}_*\sim \mathcal{N}(\hat{c}_*+\bar{\hat{r}}_*, \mathrm{var}(\hat{r}_*))$: $\hat{c}_*+\bar{\hat{r}}_*$ can be used as detection score for error detection, and $\mathrm{var}(\hat{r}_*)$ represents the uncertainty of returned detection score\\
		{\bf \hspace{-17pt}Training Phase:}
		\STATE obtain target detection score $\mathbf{c}=\{c_i=\delta_{y_i,\hat{y}_i}\}_{i=1}^N$, where $\delta_{y_i,\hat{y}_i}$ is the Kronecker delta ($\delta_{y_i,\hat{y}_i}=1$ if $y_i=\hat{y}_i$, otherwise $\delta_{y_i,\hat{y}_i}=0$)
		\STATE calculate residuals $\mathbf{r}=\{r_i = c_i - \hat{c}_i\}_{i=1}^N$
		\FOR{each optimizer step}
		\STATE calculate covariance matrix $\mathbf{K}_c((\mathcal{X},\mathbf{\bm{\sigma}}), (\mathcal{X}, \mathbf{\bm{\sigma}}))$, where each entry is given by $k_c((\mathbf{x}_i,\sigma_i), (\mathbf{x}_j,\sigma_j)) = k_\mathrm{in}(\mathbf{x}_i, \mathbf{x}_j) + k_\mathrm{out}(\sigma_i, \sigma_j),\hspace{5pt} \mathrm{for}\hspace{3pt} i,j=1,2,\ldots,N$
		\STATE optimize GP hyperparameters by maximizing log marginal likelihood $\log p(\mathbf{r}|\mathcal{X},\bm{\sigma}) = -\frac{1}{2}\mathbf{r}^\top(\mathbf{K}_c((\mathcal{X},\bm{\sigma}), (\mathcal{X}, \bm{\sigma}))+\sigma_n^2\mathbf{I})^{-1}\mathbf{r}-\frac{1}{2}\log|\mathbf{K}_c((\mathcal{X},\bm{\sigma}), (\mathcal{X}, \bm{\sigma}))+\sigma_n^2\mathbf{I}|-\frac{n}{2}\log 2\pi$
		\ENDFOR
		{\bf \hspace{-17pt}Deployment Phase:}
		\STATE calculate residual mean $\bar{\hat{r}}_* =\mathbf{k}_*^\top(\mathbf{K}_c((\mathcal{X},\bm{\sigma}), (\mathcal{X}, \bm{\sigma}))+\sigma_n^2\mathbf{I})^{-1}\mathbf{r}$ and residual variance $\mathrm{var}(\hat{r}_*) =k_c((\mathbf{x}_*,\sigma_*), (\mathbf{x}_*,\sigma_*))-\mathbf{k}_*^\top(\mathbf{K}_c((\mathcal{X},\bm{\sigma}), (\mathcal{X}, \bm{\sigma}))+\sigma_n^2\mathbf{I})^{-1}\mathbf{k}_*$, where $\mathbf{k}_*$ denotes the vector of kernel-based covariances (i.e., $k_c(\mathbf{x}_*,\mathbf{x}_i)$) between $\mathbf{x}_*$ and all training data
		\STATE return distribution of error detection score $\hat{c}^{\prime}_*\sim \mathcal{N}(\hat{c}_*+\bar{\hat{r}}_*, \mathrm{var}(\hat{r}_*))$
	\end{algorithmic}
\end{algorithm}

Figure~\ref{fig:RED_flowchart} illustrates the RED training and deployment processes conceptually, and Algorithm~\ref{alg:pseudo_RED} specifies them in detail. In the training phase, the first step is to define a target detection score $c_i$ for each training sample $(\mathbf{x}_i,y_i,\hat{y}_i,\sigma_i)$. In nature, any functions that assign target values to correct and incorrect predictions differently can be used. For simplicity, the Kronecker delta $\delta_{y_i,\hat{y}_i}$ is used in this work: all training samples that are correctly predicted by the original NN classifier receive 1 as the target detection score, and those that are incorrectly predicted receive 0. The validation dataset during the original NN training is included in the training dataset for RED. After the target detection scores are assigned, a regression problem is formulated for the mRIO model: Given the original raw features $\{\mathbf{x}_i\}_{i=1}^N$ and the corresponding softmax outputs of the original NN classifier $\{\sigma_i = [\hat{p}_{i,1}, \hat{p}_{i,2}, \dots, \hat{p}_{i,K}]\}_{i=1}^N$, predict the residuals $\mathbf{r}=\{r_i = c_i - \hat{c}_i\}_{i=1}^n$ between target detection scores $\mathbf{c}=\{c_i\}_{i=1}^N$ and the original maximum class probabilities $\mathbf{\hat{c}}=\{\hat{c}_i=\mathrm{max}(\sigma_i)\}_{i=1}^N$.

The mRIO model relies on an I/O kernel consisting of two components: the input kernel $k_\mathrm{in}(\mathbf{x}_i, \mathbf{x}_j)$, which measures covariances in the raw feature space, and the modified multi-output kernel $k_\mathrm{out}(\sigma_i, \sigma_j)$, which calculates covariances in the softmax output space. The hyperparameters of the I/O kernel are optimized to maximize the log marginal likelihood $\log p(\mathbf{r}|\mathcal{X},\bm{\sigma})$. In the deployment phase, given a new data point $\mathbf{x}_*$, the trained mRIO model provides a Gaussian distribution for the estimated residual $\hat{r}_*\sim \mathcal{N}(\bar{\hat{r}}_*, \mathrm{var}(\hat{r}_*))$. By adding the estimated residual back to the original maximum class probability $\hat{c}_*$, a distribution of detection score is obtained as $\hat{c}^{\prime}_*\sim \mathcal{N}(\hat{c}_*+\bar{\hat{r}}_*, \mathrm{var}(\hat{r}_*))$. The mean $\hat{c}_*+\bar{\hat{r}}_*$ can be directly used as a quantitative metric for error detection, and the variance $\mathrm{var}(\hat{r}_*)$ represents the corresponding uncertainty of the detection score.	
\section{Empirical Evaluation}
\label{sec:empirical_study}
In this section, the error detection performance of RED is evaluated comprehensively on 125 UCI datasets, comparing it to other related methods. Its generality is then evaluated by applying it to two other base models, and its scale-up properties measured in two larger deep learning architectures solving two vision tasks. Finally, RED's potential to improve robustness more broadly is demonstrated in an empirical study involving OOD and adversarial samples.

\subsection{Comparisons with Related Approaches}
\label{subsec:exp_UCI}
As a comprehensive evaluation of RED, an empirical comparison with seven related approaches on 125 UCI datasets \citep{Dua2017} was performed. These approaches include maximum class probability (MCP) baseline \citep{Hendrycks2017}, three state-of-the-art approaches, namely Trust Score \citep[T-Score;][]{Jiang18}, ConfidNet \citep[C-net;][]{Charles19}, and Introspection-Net \citep[I-net;][]{Jonathan19}, as well as three earlier approaches, i.e.\ entropy of the original softmax outputs \citep{Steinhardt2016}, DNGO \citep{Snoek2015}, and the original SVGP \citep{Hensman2013,Hensman2015}. The 125 UCI datasets include 121 datasets used by \citet{Klambauer17} and four more recent ones. Preliminary tests indicate that RED is not sensitive to the choice of kernel functions, so the standard RBF (Radial Basis Function) kernel is used in the current RED implementation. See Appendix for full details about the datasets and parametric setup of all tested algorithms. Source codes are provided in \href{https://github.com/cognizant-ai-labs/red-paper}{https://github.com/cognizant-ai-labs/red-paper}.

Following the experimental setup of \citet{Hendrycks2017, Charles19, Jonathan19}, the task for each algorithm is to provide a detection score for each testing point. An error detector can then use a predefined fixed threshold on this score to decide which points are probably misclassified by the original NN classifier. For RED, the mean $\hat{c}_*+\bar{\hat{r}}_*$ was used as the reported detection score. Five threshold-independent performance metrics were used to compare the methods: AUPR-Error, which computes the area under the Precision-Recall (AUPR) Curve when treating incorrect predictions as positive class during the detection; AUPR-Success, which is similar to AUPR-Error but uses correct predictions as positive class; AUROC, which computes the area under receiver operating characteristic (ROC) curve for the error detection task; AP-Error, which computes the average precision (AP) under different thresholds treating incorrect predictions as the positive class; and AP-Success, which is similar to AP-Error but uses correct predictions as the positive class. AUPR and AUROC are commonly used in prior work \citep{Hendrycks2017, Charles19, Jonathan19}, but as discussed by \citet{Davis06} and \citet{Flach15}, AUPR may provide overly-optimistic measurement of performance. Moreover, AUROC is sometimes less informative \cite{Manning1999} and not ideal when the positive class and negative class have greatly differing base rates \cite{Hendrycks2017} (this happens when the base classifier has high prediction accuracy so there are only few misclassified examples). To compensate for these issues, AP-Error and AP-Success are included as additional metrics. Since the target of all tested approaches is to detect misclassification errors, the following discussion will focus more on AP-Error and AUPR-Error.
\begin{table}[t]
	\scriptsize
	\centering
	\setlength{\tabcolsep}{3.6pt}
	\begin{tabular}{l c c c c c}
		\toprule
		\multirow{2}{*}{Method} & AP-Error & AUPR-Error & AP-Success & AUPR-Success & AUROC \\ 
		{}	& mean$\pm$std & mean$\pm$std & mean$\pm$std & mean$\pm$std & mean$\pm$std \\
		\hline
		RED & \textbf{1.39$\pm$0.61}* & \textbf{1.49$\pm$0.78}* & \textbf{1.74$\pm$0.97}*  & \textbf{1.80$\pm$1.03}* & \textbf{1.65$\pm$0.82}* \\
		MCP & 2.93$\pm$ 0.89 & 3.06$\pm$0.92 & 2.77$\pm$1.07  & 2.75$\pm$1.11 & 2.80$\pm$1.08 \\
		T-Score & 3.92$\pm$2.45 & 3.86$\pm$2.50 & 3.64$\pm$2.25  & 3.61$\pm$2.25 & 3.76$\pm$2.31 \\
		C-Net & 6.13$\pm$1.37 & 6.33$\pm$1.38 & 6.07$\pm$1.51 & 6.07$\pm$1.41 & 5.97$\pm$1.45 \\
		I-Net & 5.34$\pm$1.65 & 5.38$\pm$1.65 & 5.83$\pm$1.46 & 5.89$\pm$1.51 & 5.71$\pm$1.50  \\
		Entropy & 3.47$\pm$1.08 & 3.59$\pm$1.19 & 3.19$\pm$1.26 & 3.23$\pm$1.32 & 3.26$\pm$1.28  \\
		DNGO & 6.19$\pm$1.51 & 5.46$\pm$1.82 & 6.84$\pm$1.33 & 6.80$\pm$1.44  & 6.57$\pm$1.47 \\
		SVGP & 6.59$\pm$1.60 & 6.80$\pm$1.49 & 5.89$\pm$1.54  & 5.83$\pm$1.49 & 6.24$\pm$1.61 \\
		\bottomrule
	\end{tabular}\\
	\tiny{The symbol * indicates that the differences between the marked entry and all other counterparts are statistically significant at the 5\% significance level for both paired $t$-test and Wilcoxon test. The best entries that are significantly better than all the others under both tests are in boldface.}\\
	\caption{\label{tab:mean_rank_UCI} Mean rank on UCI datasets}
\end{table}
\begin{figure}[ht]
	\centering
	\hspace{10pt}\includegraphics[width=0.83\linewidth]{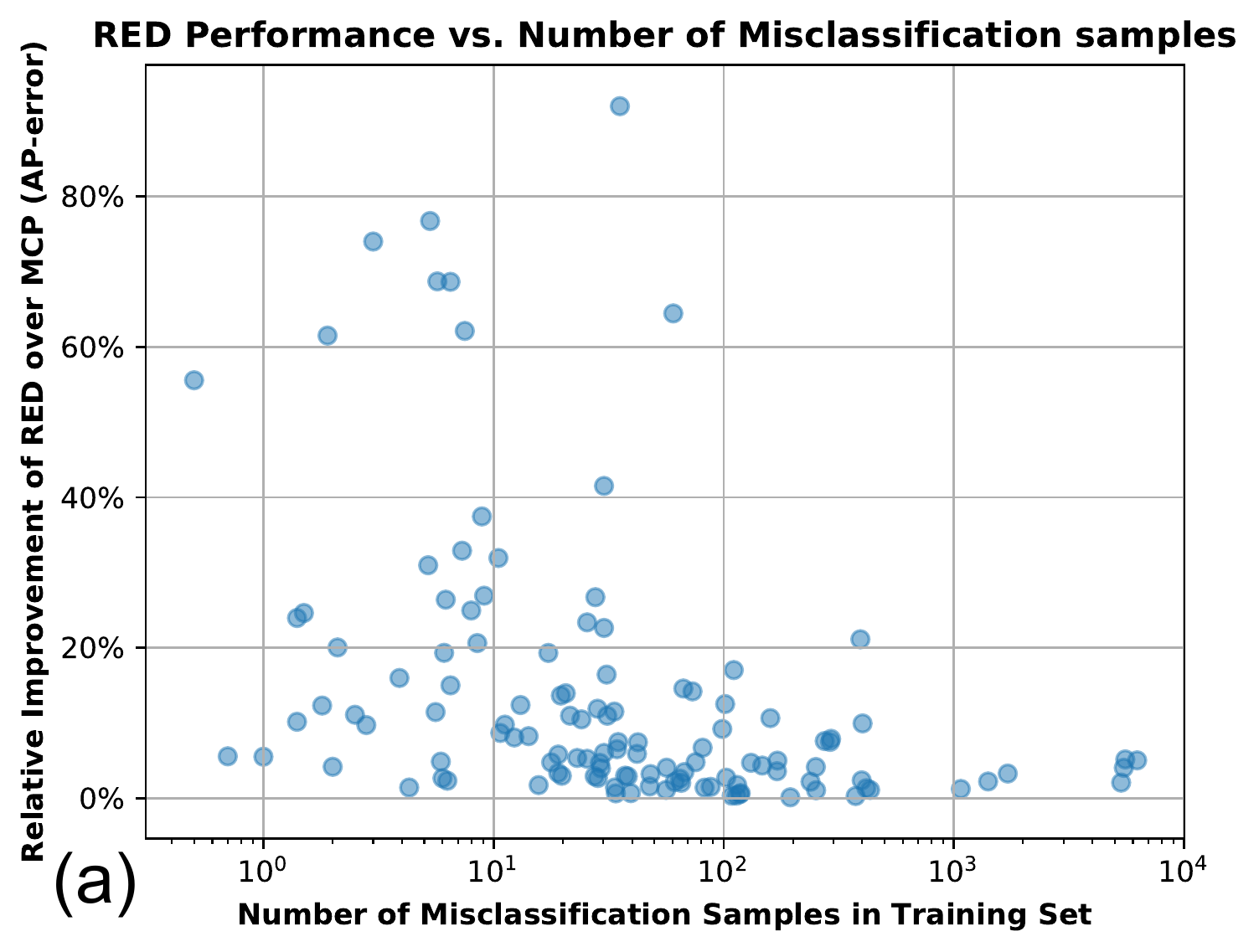}\\
	\includegraphics[width=0.81\linewidth]{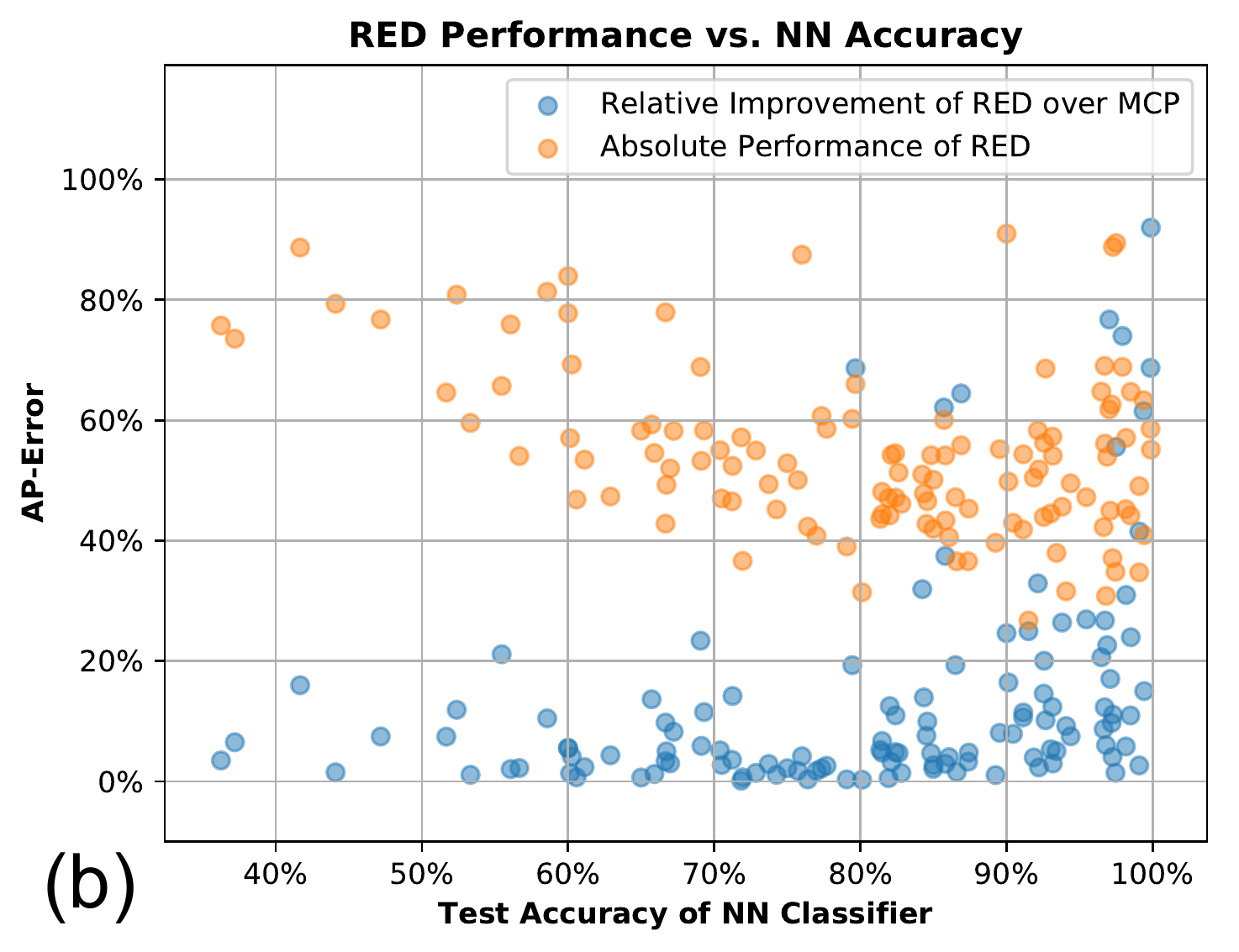}\\
	\caption{\textbf{Impact of Number of Misclassification Samples (a) and Impact of Classifier Accuracy (b)}.  \label{fig:misclassification_impact}
		Each dot represents one dataset, averaged over 10 independent runs. In some runs of a dataset, there are no misclassified samples (100\% accuracy), so the average number of misclassified samples may be less than 1. The improvement of RED over MCP baseline is more significant for situations where number of misclassification samples is small (few-shot learning) or accuracies of base NN classifiers are high (imbalanced training set), indicating that RED provides the largest advantage in extreme cases..
	}
\end{figure}
\begin{table}[t]
	\scriptsize
	\centering
	\setlength{\tabcolsep}{3.6pt}
	\begin{tabular}{l c c c c c}
		\toprule
		RED & AP-Error & AUPR-Error & AP-Success & AUPR-Success & AUROC \\ 
		vs.	& + / = / - & + / = / - & + / = / - & + / = / - & + / = / - \\
		\hline
		MCP & 87 / 35 / 0 & 90 / 32 / 0 & 58 / 63 / 1 & 56 / 65 / 1 & 61 / 60 / 1 \\
		T-Score & 53 / 44 / 16 & 49 / 47 / 17 & 50 / 47 / 16 & 48 / 49 / 16 & 59 / 37 / 17 \\
		C-Net & 100 / 22 / 0 & 100 / 22 / 0 & 106 / 16 / 0 & 106 / 16 / 0 & 109 / 13 / 0 \\
		I-Net & 93 / 29 / 0 & 90 / 32 / 0 & 98 / 24 / 0 & 98 / 24 / 0 & 101 / 21 / 0 \\
		Entropy & 74 / 47 / 1 & 75 / 46 / 1 & 53 / 68 / 1 & 53 / 68 / 1 & 52 / 69 / 1 \\
		DNGO & 92 / 17 / 0 & 73 / 31 / 5 & 99 / 10 / 0 & 97 / 12 / 0 & 98 / 11 / 0 \\
		SVGP & 98 / 23 / 1 & 98 / 23 / 1 & 97 / 25 / 0 & 97 / 25 / 0 & 102 / 19 / 1 \\
		\hline
		BNN-M & 102 / 20 / 0 & 104 / 18 / 0 & 95 / 26 / 1 & 88 / 33 / 1 & 95 / 26 / 1 \\
		BNN-E & 67 / 53 / 2 & 68 / 52 / 2 & 48 / 66 / 8 & 48 / 66 / 8 & 53 / 64 / 5 \\
		MCD-M & 87 / 35 / 0 & 88 / 34 / 0 & 70 / 52 / 0 & 67 / 55 / 0 & 71 / 51 / 0 \\
		MCD-E & 54 / 68 / 0 & 55 / 67 / 0 & 38 / 77 / 7 & 38 / 76 / 8 & 42 / 74 / 6 \\
		\hline
		BLR-res & 77 / 43 / 0 & 76 / 44 / 0 & 92 / 28 / 0 & 90 / 30 / 0 & 88 / 32 / 0 \\
		\bottomrule
	\end{tabular}\\
	\tiny{The columns labeled + show the number of datasets on which RED performs significantly better at the 5\% significance level in a paired $t$-test, Wilcoxon test, or both; those labeled - represent the contrary case; those labeled = represent no statistical significance.}\\
	\caption{\label{tab:pairwise_comparison_UCI} A pairwise comparison between RED and other methods on UCI datasets}
\end{table}
\begin{figure*}[t]
	\centering
	\includegraphics[width=0.19\linewidth]{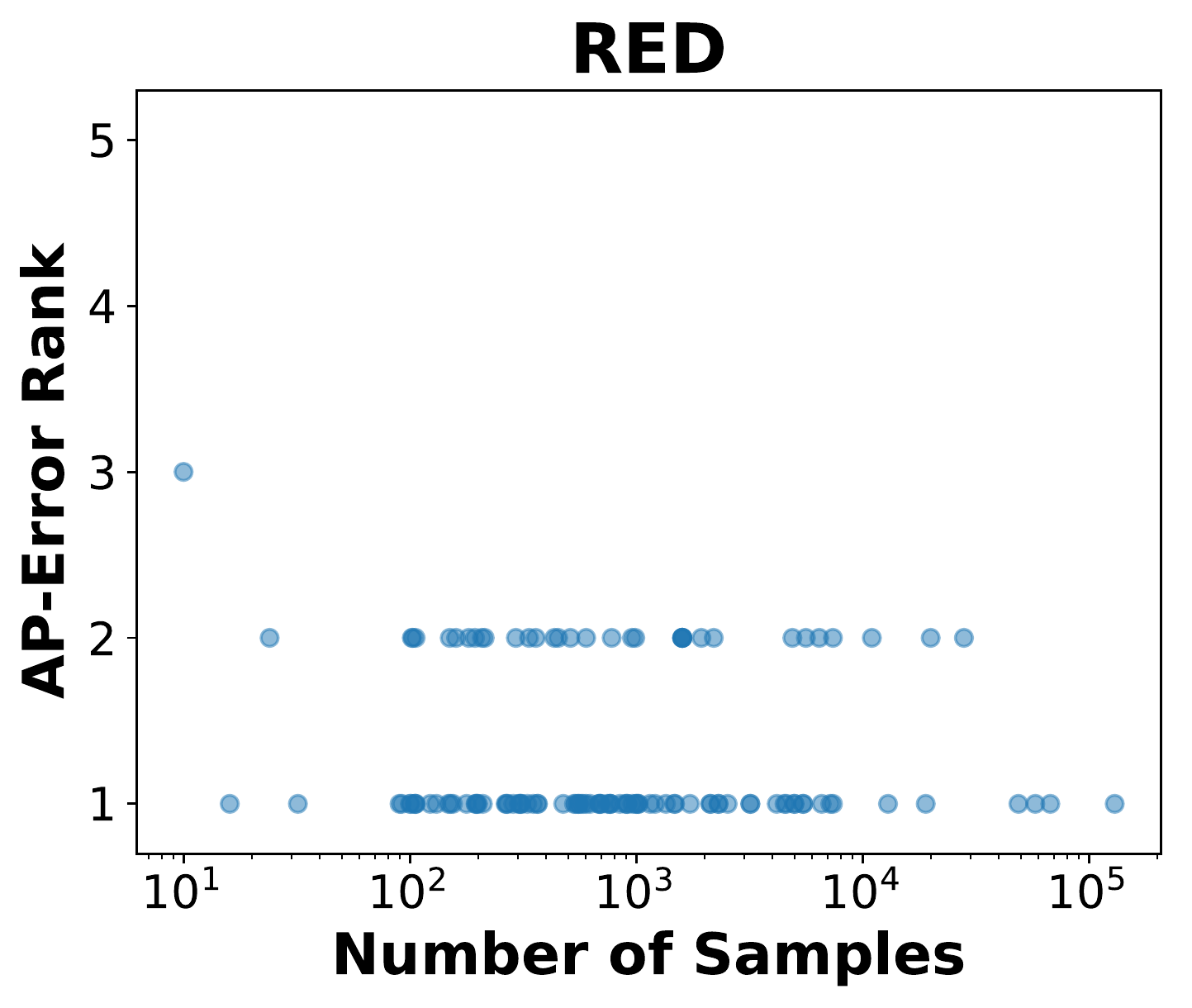}
	\includegraphics[width=0.19\linewidth]{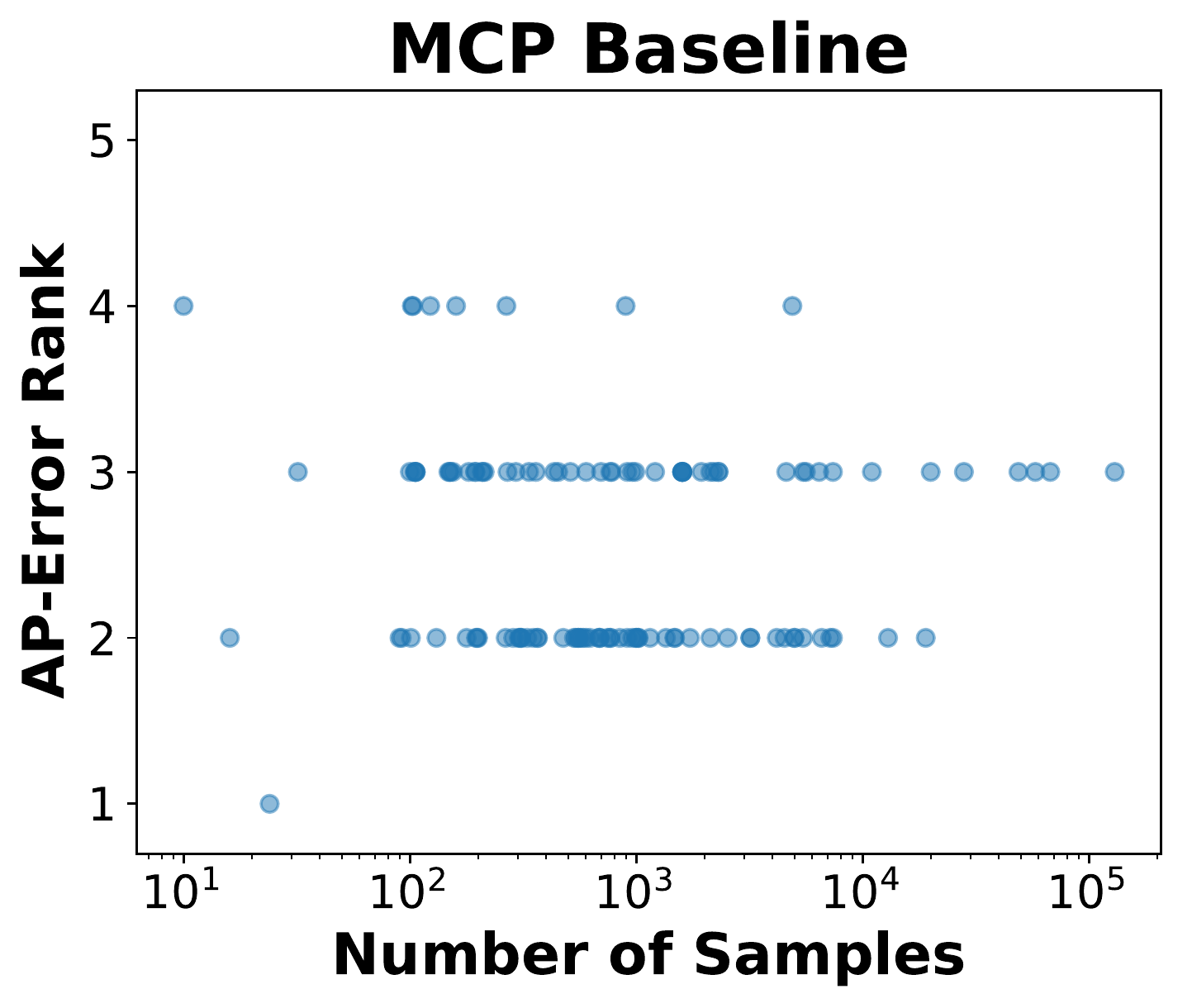}
	\includegraphics[width=0.19\linewidth]{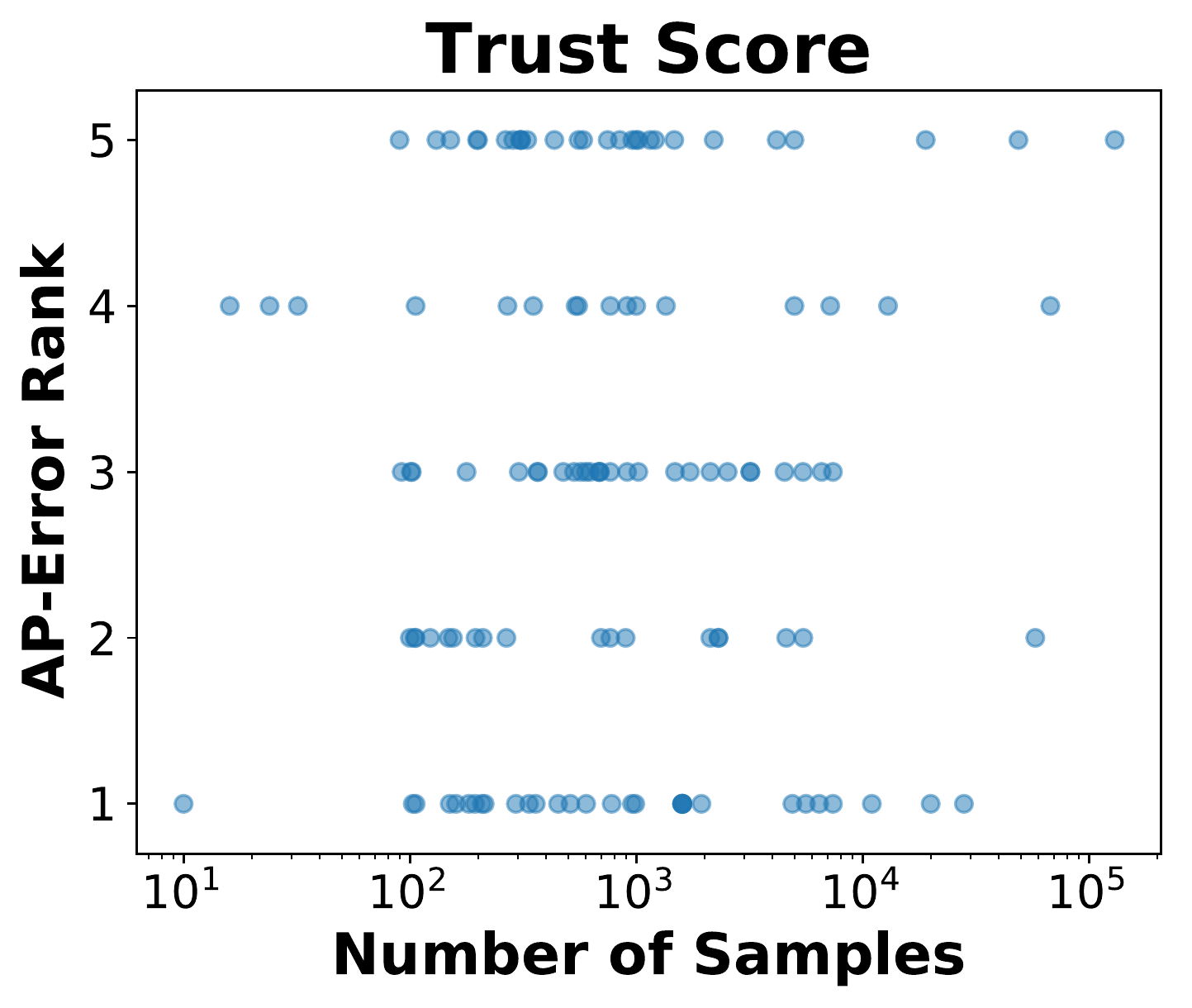}
	\includegraphics[width=0.19\linewidth]{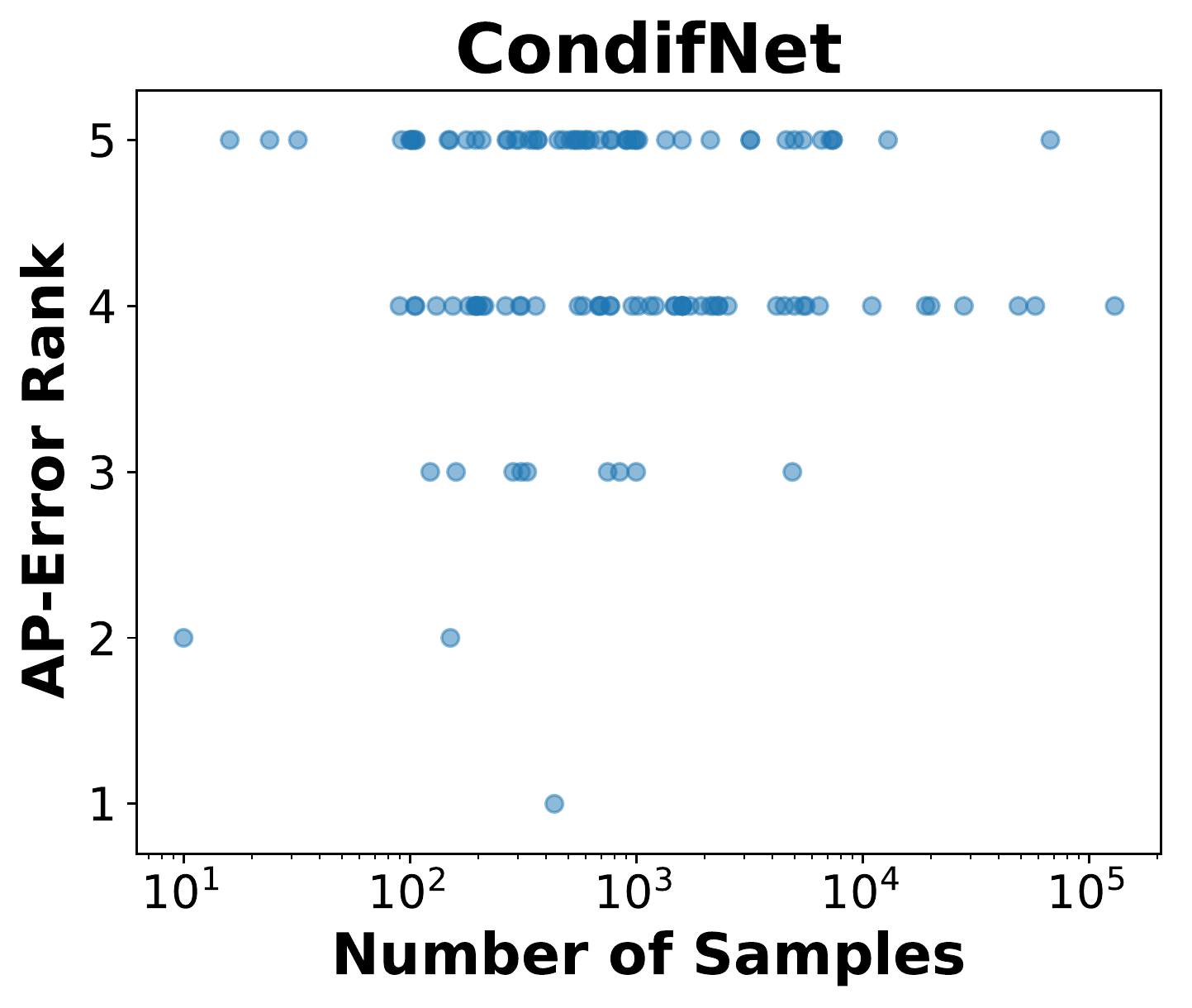}
	\includegraphics[width=0.19\linewidth]{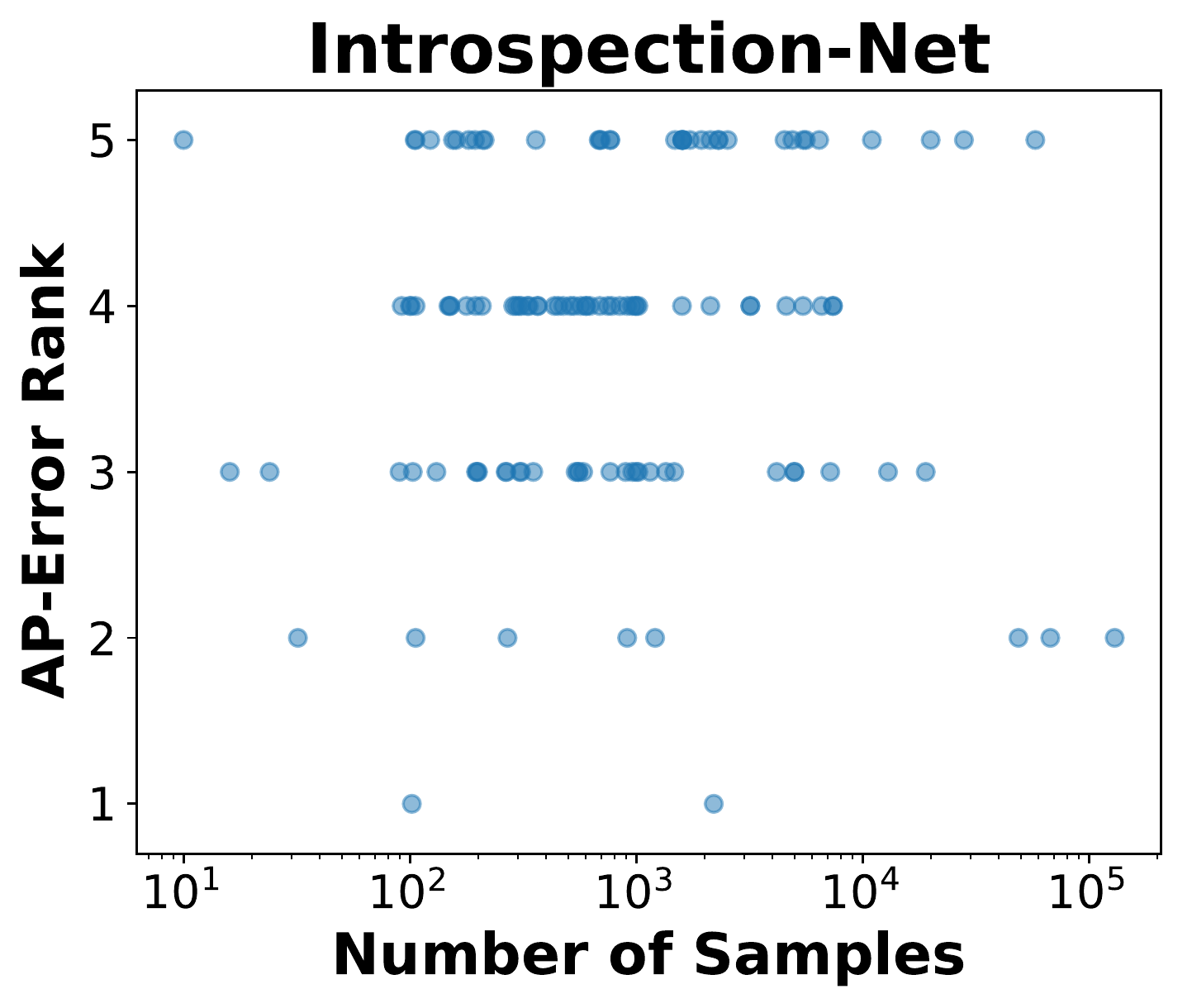}
	\includegraphics[width=0.19\linewidth]{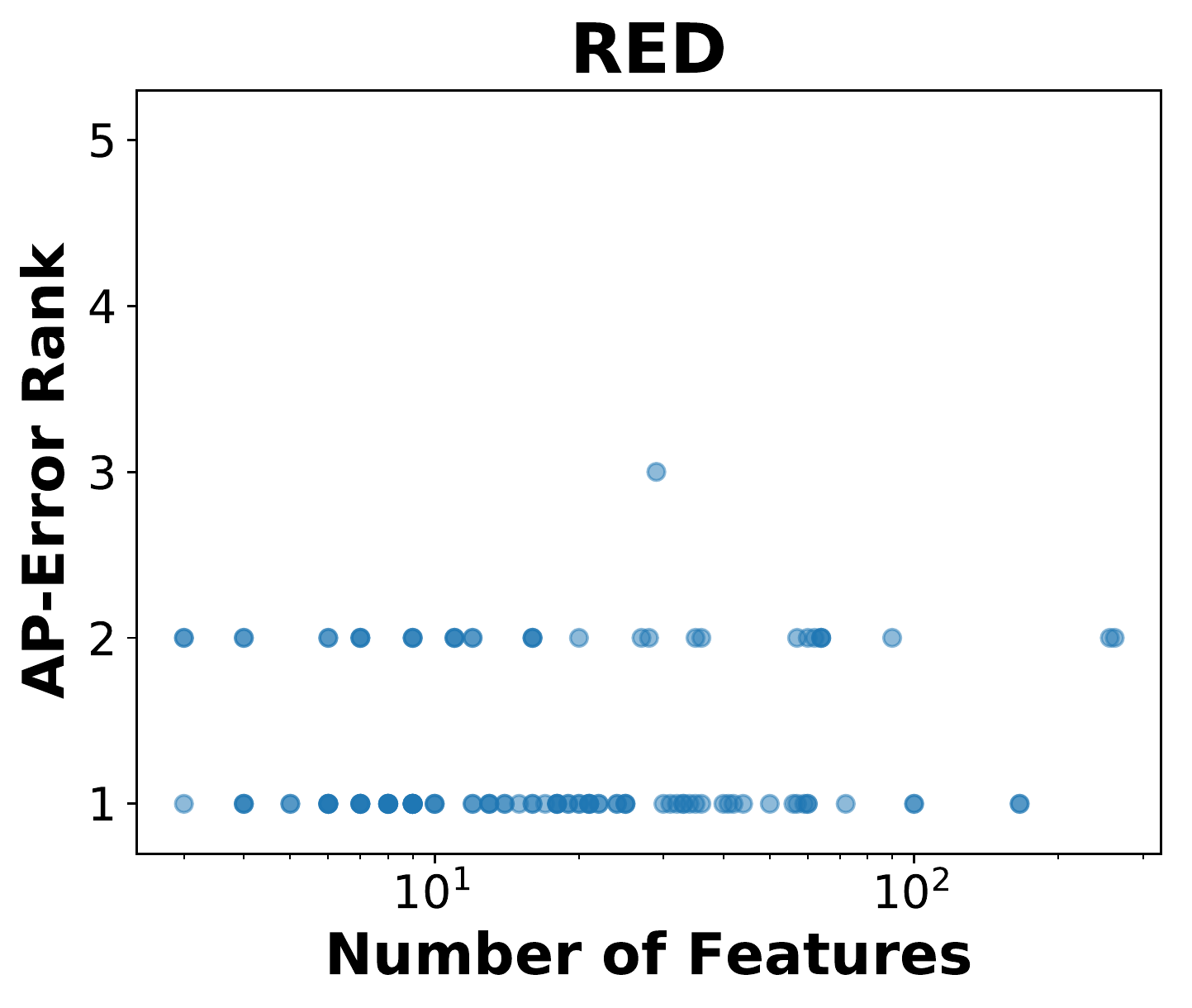}
	\includegraphics[width=0.19\linewidth]{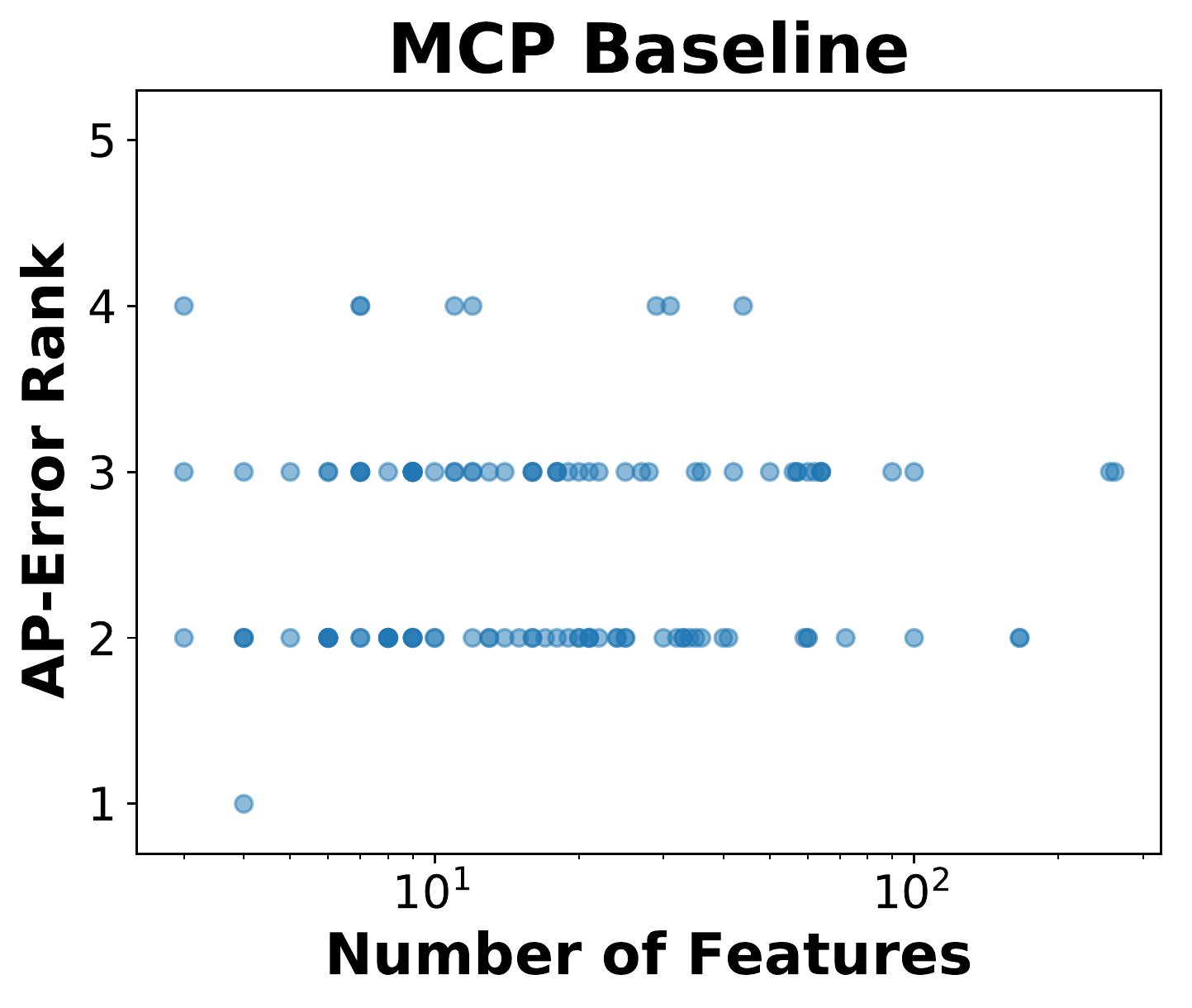}
	\includegraphics[width=0.19\linewidth]{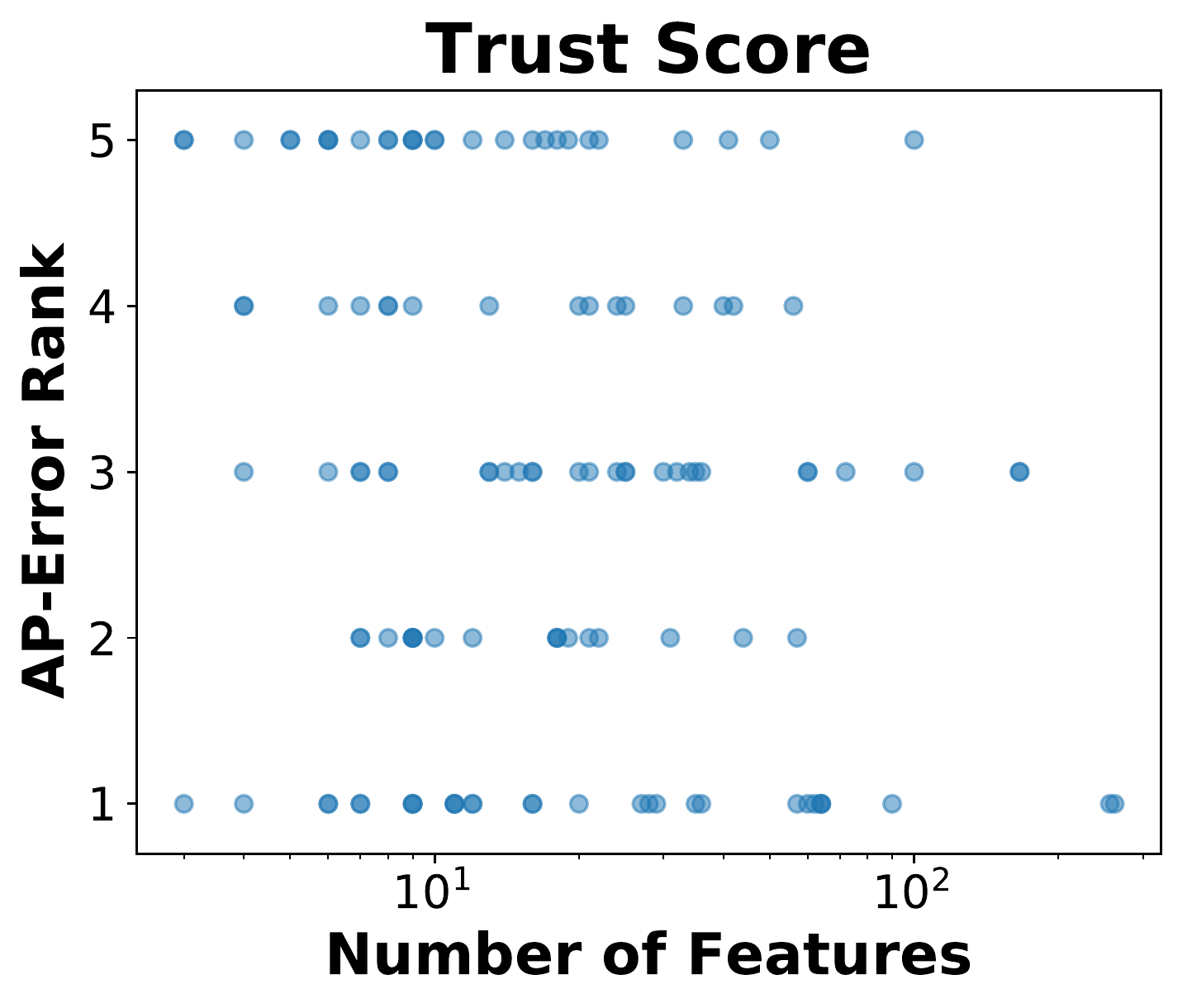}
	\includegraphics[width=0.19\linewidth]{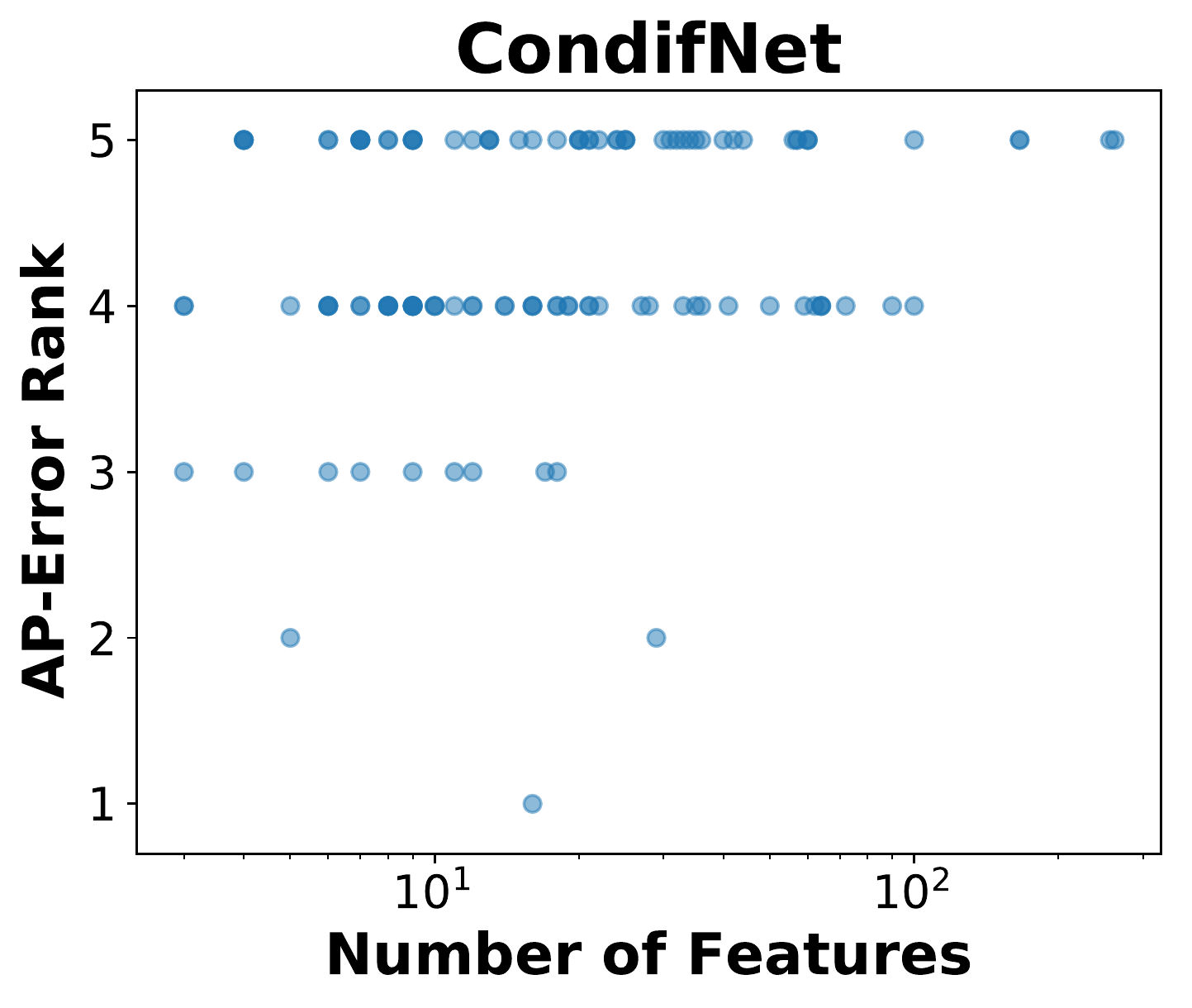}
	\includegraphics[width=0.19\linewidth]{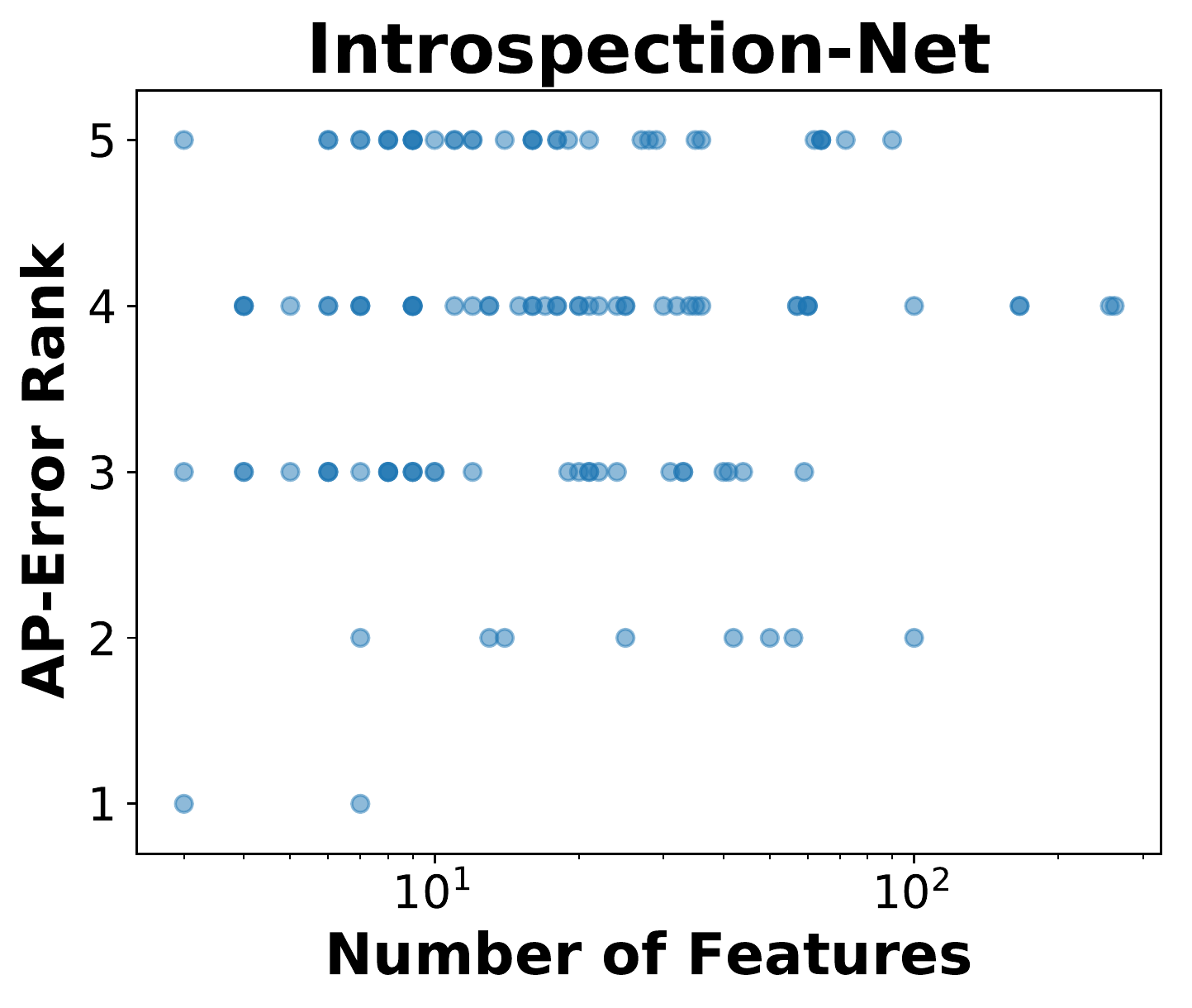}
	\caption{\textbf{Performance ranks across dataset sizes and feature dimensionalities on UCI datasets.} \label{fig:rank_dataset_info}
		Each plot represents the distribution of relative ranks for one method (each column) as a function of the dataset size (top row) and the feature dimensionality (bottom row). Each dot in each plot represents the relative rank in one dataset. RED performs  consistently well over datasets of different sizes and feature dimensionalities. Trust Score performs inconsistently, and ConfidNet performs poorly on larger datasets.
	}
\end{figure*}

Ten independent runs were conducted for each dataset. During each run, the dataset was randomly split into training dataset and testing dataset, and a standard NN classifier trained and evaluated on them. The same dataset split and trained NN classifier were used to evaluate all methods. The testbed covers a diversified collection of dataset sizes and classifier accuracies, thereby enabling a comprehensive evaluation of the tested approaches. Full details of the experimental setup are provided in Appendix~\ref{subsec:setup_UCI125}.

Table~\ref{tab:mean_rank_UCI} shows the ranks of each algorithm averaged over all 125 UCI datasets (see Appendix~\ref{subsec:detailed_results_UCI125} for detailed results). The rank of each algorithm on each dataset is based on the average performance over 10 independent runs. RED performs best on all metrics; the performance differences between RED and all other methods are statistically significant under paired $t$-test and Wilcoxon test. Trust Score has the highest standard deviation, suggesting that its performance varies significantly across different datasets.

Table~\ref{tab:pairwise_comparison_UCI} shows how often RED performs statistically significantly better, how often the performance is not significantly different, and how often it performs significantly worse than the other methods. RED is most often significantly better, and very rarely worse. In a handful of datasets Trust Score is better, but most often it is not.

To illustrate the impact of misclassification samples in training set, Figure~\ref{fig:misclassification_impact} shows the performance of RED under different numbers of misclassification samples and base NN classifier accuracies. RED consistently improves over the MCP baseline, and this improvement is more significant for extreme cases, i.e. when only a limited number of misclassification samples is provided (i.e. in case of few-shot learning) or the base NN classifier has high accuracy (i.e. in case of an imbalanced training set), demonstrating the robustness of RED. RED's AP-Error is generally higher with low-accuracy classifiers. This is because higher ratio of misclassification samples naturally leads to higher detection precision, and an accuracy around 50\% leads to a more balanced training set for the error detector.

To further study the robustness of RED compared to the baseline and the three state-of-the-art approaches, Figure~\ref{fig:rank_dataset_info} shows the distribution of the relative rank of RED, MCP baseline, Trust Score, ConfidNet and Introspection-Net as a function of the number of samples and the number of features in the dataset. These plots are based on the AP-Error metric; other metrics provide similar results. RED performs consistently well over different dataset sizes and feature dimensionalities. Trust Score performs best in several datasets, but occasionally also worst in both small and large datasets, making it a rather unreliable choice. ConfidNet generally exhibits worse performance on datasets with large dataset sizes and high feature dimensionalities, i.e.\ it does not scale well to larger problems.


To evaluate whether GP is indeed an appropriat model for the RED framework, it was replaced by a Bayesian linear regressor \citep[BLR;][]{Snoek2015}, with all other components unchanged. This BLR-residual (BLR-res) variant was then compared with the original RED in all 125 UCI datasets (see Appendix~\ref{subsec:setup_UCI125} for the setup). Results in Table~\ref{tab:pairwise_comparison_UCI} (last row) show that RED dominates BLR-res, indicating that GP is a good choice for error detection tasks.


\subsection{Generality wrt.\ Base Models}
\label{subsec:genbase}

To evaluate generality of RED, it was applied to two other base models: an NN classifier using Monte Carlo-dropout (MCD) technique \citep{Gal2016} and a Bayesian Neural Network (BNN) classifier \citep{wen2018}. They were each trained as base classifiers, and RED was then applied to each of them (implementation details are provided in Appendix~\ref{subsec:setup_UCI125}). Experiments analogous to those in Section~\ref{subsec:exp_UCI} were performed on 125 UCI datasets in both cases. Table~\ref{tab:pairwise_comparison_UCI} (rows starting with "BNN" or "MCD") summarizes the pairwise comparisons between RED and the internal detection scores returned by the base models (see Appendix~\ref{subsec:detailed_results_UCI125} for detailed results). "-M" and "-E" represent the maximum class probability and entropy of softmax outputs, respectively, after averaging over 100 test-time samplings. RED significantly improves MCD and BNN classifier in most datasets, demonstrating that it is a general technique that can be applied to a variety of models.
\begin{table*}[t]
	\centering
	\scriptsize
	\begin{tabular}{c c c c c c c c c c}
		\toprule
		\multirow{2}{*}{Task} &\multirow{2}{*}{Metric} & RED & MCP Baseline & Trust Score & ConfidNet & Introspection-Net & Entropy & DNGO & SVGP \\
		{} &{}	& mean$\pm$std & mean$\pm$std & mean$\pm$std & mean$\pm$std & mean$\pm$std & mean$\pm$std & mean$\pm$std & mean$\pm$std\\
		\hline
		VGG16 on &AP-Error(\%) & \textbf{49.88$\pm$1.99}* & 47.09$\pm$2.19 & 48.76$\pm$2.28 & 45.80$\pm$2.85 & 42.11$\pm$1.98 & 47.91$\pm$2.17 & 33.91$\pm$2.94 & 40.71$\pm$2.33\\
		CIFAR-10&AUPR-Error(\%) & \textbf{49.79$\pm$2.00}* & 46.99$\pm$2.21 & 48.68$\pm$2.29 & 45.70$\pm$2.86 & 42.01$\pm$1.98 & 47.81$\pm$2.19 & 34.43$\pm$2.92 & 40.60$\pm$2.34\\
		\hline
		WRN-10-4 on &AP-Error(\%) & \textbf{52.51$\pm$2.81}* & 48.76$\pm$1.47 & 51.15$\pm$3.29 & 48.34$\pm$3.05 & 40.52$\pm$3.64 & 48.12$\pm$1.51 & 28.96$\pm$3.54 & 6.43$\pm$0.32\\
		CIFAR-10&AUPR-Error(\%) & \textbf{52.45$\pm$2.82}* & 48.70$\pm$1.48 & 51.09$\pm$3.30 & 48.27$\pm$3.05 & 40.44$\pm$3.65 & 48.06$\pm$1.51 & 29.40$\pm$3.60 & 6.42$\pm$0.32\\
		\hline
		VGG19 on &AP-Error(\%) & \textbf{73.40$\pm$1.05}* & 71.78$\pm$1.24 & 72.63$\pm$1.20 & 72.12$\pm$1.33 & 69.73$\pm$1.13 & 72.95$\pm$1.19 & 55.44$\pm$1.64 & 29.24$\pm$0.90\\
		CIFAR-100&AUPR-Error(\%) & \textbf{73.39$\pm$1.05}* & 71.77$\pm$1.25 & 72.62$\pm$1.20 & 72.10$\pm$1.13 & 69.72$\pm$1.13 & 72.94$\pm$1.20 & 55.44$\pm$1.64 & 33.64$\pm$1.24\\
		\bottomrule
	\end{tabular}\\
	\tiny{The symbol * indicates that the differences between the marked entry and all other counterparts are statistically significant at the 5\% significance level for both paired $t$-test and Wilcoxon test. The best entries that are significantly better than all the others under both tests are in boldface.}\\
	\caption{\label{tab:results_VGG16_CIFAR10} Error detection performance with deep NN models on CIFAR-10 and CIFAR-100}
\end{table*}
\subsection{Scaling up to Larger Architectures}
\label{subsec:VGG16_CIFAR10}

To confirm that the RED approach scales up to large deep learning architectures, a VGG16 model \citep{Simonyan15} and a Wide Residual Network (WRN) model \citep{ZagoruykoK2016} was trained on the CIFAR-10 dataset \citep{Krizhevsky2009}, and a VGG19 model \citep{Simonyan15} was trained on the CIFAR-100 dataset \citep{Krizhevsky2009}, using state-of-the-art training pipelines \citep{Bingham2020} (see Appendix~\ref{subsec:setup_VGG16_CIFAR10} for details). In order to remove the influence of feature extraction in image preprocessing and to make the comparison fair, all approaches used the same logit outputs of the trained VGG/WRN model as their input features. 10 independent runs are performed. During each run, a VGG/WRN model is trained, and all the methods are evaluated based on this VGG/WRN model.

Table~\ref{tab:results_VGG16_CIFAR10} shows the results on the two main error detection performance metrics (note that the table lists absolute values instead of rankings along each metric). Trust Score performs much better than in previous literatures \citep{Charles19}. This difference may be due to the fact that logit outputs are used as input features here, whereas \citet{Charles19} utilized a higher dimensional feature space for Trust Score. RED significantly outperforms all the counterparts in both metrics. This result demonstrates the advantages of RED in scaling up to larger architectures.

\subsection{Distinguishing In-sample Errors from OOD and Adversarial Samples}
\label{subsec:OOD_adversarial}

In all experiments so far, the mean of detection score $\hat{c}_*+\bar{\hat{r}}_*$ was used as RED's detection metric. Although good performance was observed in error detection by only using the mean, the variance of detection score $\mathrm{var}(\hat{r}_*)$ may be helpful if the scenario is more complex, e.g., the dataset includes some OOD data, or even adversarial data.
\begin{figure*}[t]
	\centering
	\includegraphics[width=0.32\linewidth]{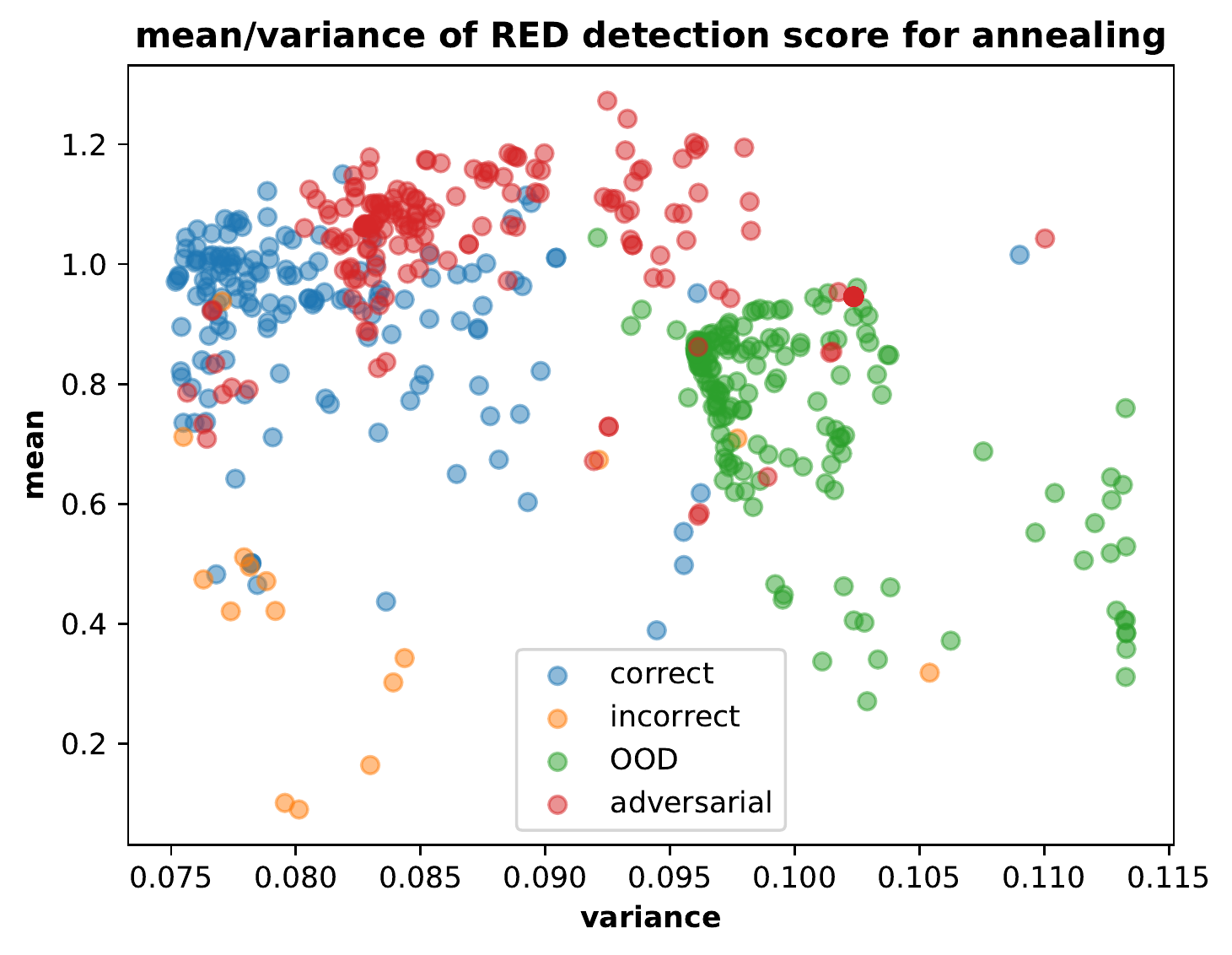}
	\includegraphics[width=0.31\linewidth]{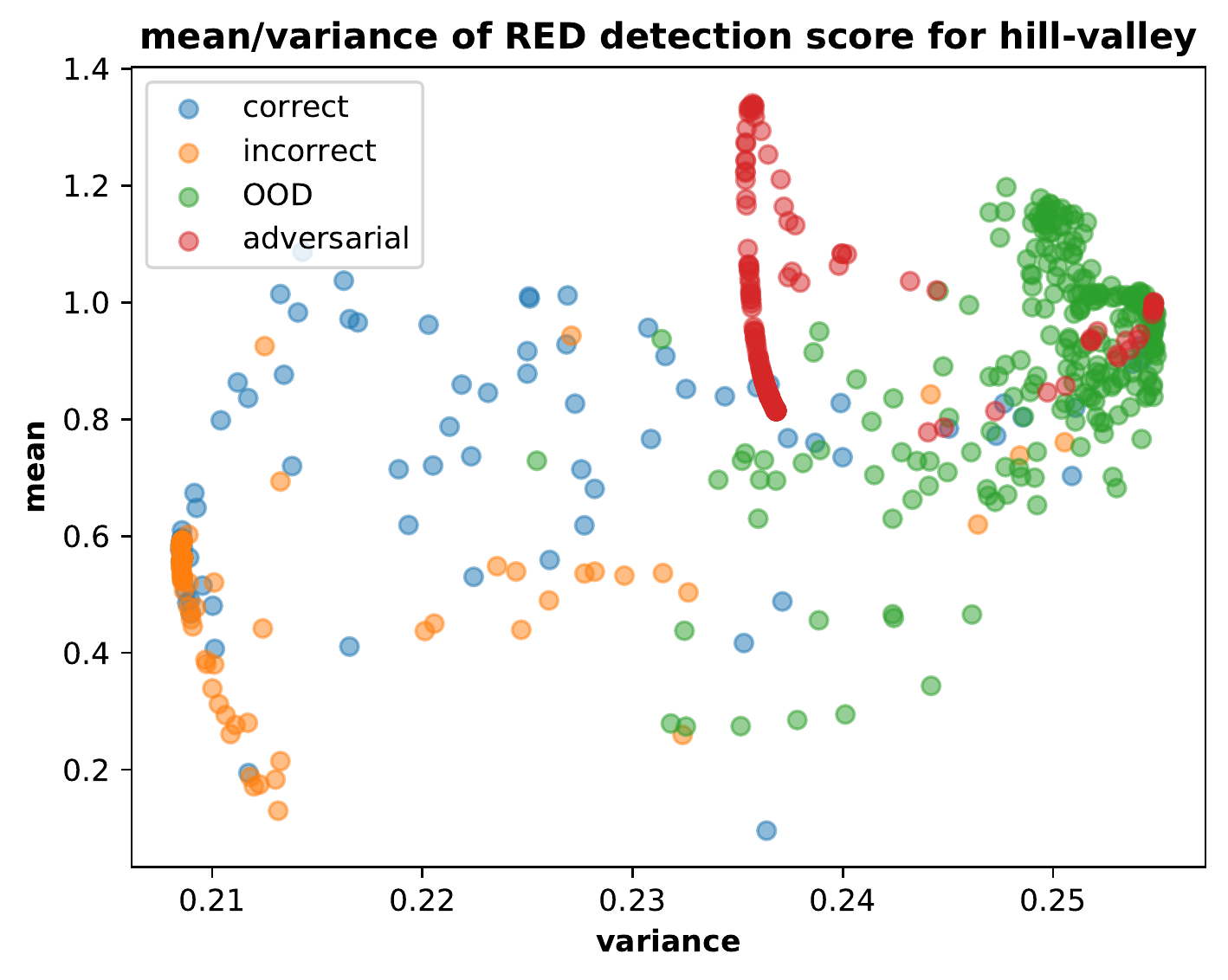}
	\includegraphics[width=0.348\linewidth]{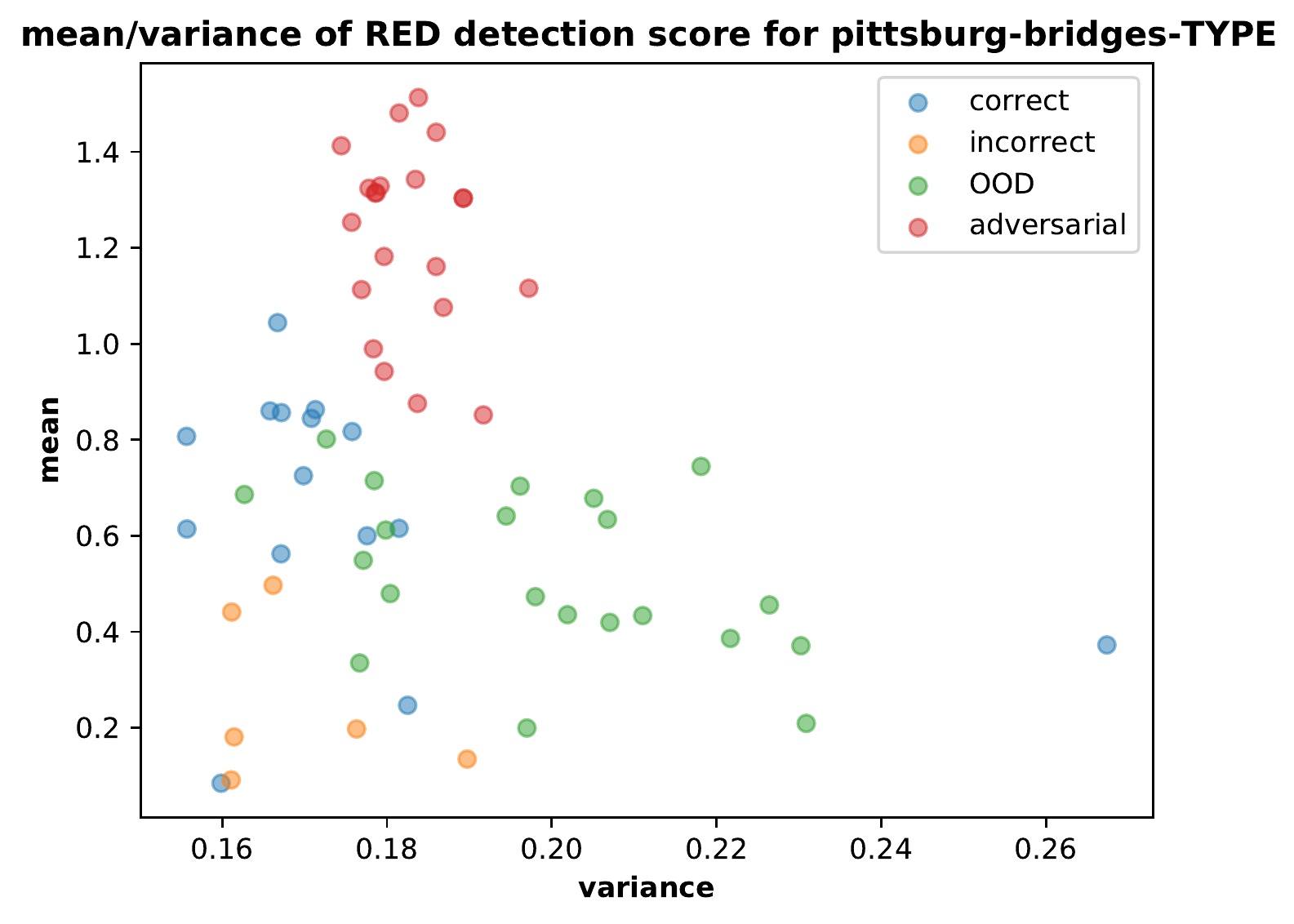}\\
	\caption{\textbf{Distinguishing in-sample errors from OOD and adversarial samples based on mean and variance of detection scores.} \label{fig:OOD_adversarial}
		Each dot represents one sample in the testing set in the corresponding UCI task. The horizontal axis denotes the variance of RED-returned detection score, and the vertical axis denotes the mean.  If an in-distribution sample is correctly classified by original NN classifier, it is marked as "correct", otherwise it is marked "incorrect". Mean is a good separator of correct and incorrect classifications. High variance, on the other hand, indicates that RED is uncertain about its detection score, which can be used to identify OOD and adversarial samples.
	}
\end{figure*}
\begin{table}[t]
	\scriptsize
	\centering
	\setlength{\tabcolsep}{3.6pt}
	\begin{tabular}{l c c c c }
		\toprule
		RED-variance & AP-OOD & AUPR-OOD & AP-Adversarial & AUPR-Adversarial \\ 
		vs.	& + / = / - & + / = / - & + / = / - & + / = / - \\
		\hline
		MCP baseline& 101 / 15 / 9 & 101 / 13 / 11 & 122 / 3 / 0 & 124 / 1 / 0\\
		RED-mean & 100 / 14 / 11 & 100 / 13 / 12 & 122 / 3 / 0 & 122 / 3 / 0 \\
		\bottomrule
	\end{tabular}\\
	\tiny{The columns labeled + show the number of datasets on which RED performs significantly better at the 5\% significance level in a paired $t$-test, Wilcoxon test, or both; those labeled - represent the contrary case; those labeled = represent no statistical significance.}\\
	\caption{\label{tab:pairwise_comparison_OOD+adv} A pairwise comparison between RED-variance and other methods on detection of OOD and adversarial samples}
\end{table}
\begin{table}[t]
	\centering
	\scriptsize
	\begin{tabular}{c c}
		\toprule
		AP-OOD(\%) & AUPR-OOD(\%) \\
		RED-variance / MCP baseline & RED-variance / MCP baseline\\
		\hline
		\textbf{86.282$\pm$2.212}* / 82.964$\pm$1.850 & \textbf{86.276$\pm$2.213}* / 82.958$\pm$1.851\\
		\bottomrule
	\end{tabular}\\
	\tiny{The symbol * indicates that the difference between the RED-variance and the MCP baseline is statistically significant at the 5\% significance level for both paired $t$-test and Wilcoxon test. The significantly better entries under both tests are in boldface.}\\
	\caption{\label{tab:results_CIFAR10vsSVHN} Performance on detecting OOD samples (SVHN data) from CIFAR-10 data (mean$\pm$std over 10 runs)}
\end{table}

RED was evaluated in such a scenario by manually adding OOD and adversarial data into the test set of all 125 UCI datasets. The synthetic OOD and adversarial samples were created to be highly deceptive, aiming to evaluate the performance of RED under difficult circumstances. The OOD data were sampled from a Gaussian distribution with mean 0 and variance 1. All samples from original dataset were normalized to have mean 0 and variance 1 for each feature dimension so that the OOD data and in-distribution data had similar scales. The adversarial data simulate situations where negligible modifications to training samples cause the original NN classifier to predict incorrectly with highest confidence \citep{Goodfellow14}. Full setup details are provided in Appendix~\ref{subsec:setup_OOD_adv}.

Figure~\ref{fig:OOD_adversarial} shows the distribution of mean and variance of detection scores for testing samples, including correctly and incorrectly labeled actual samples, as well as the synthetic OOD and adversarial samples. The mean is a good separator for correctly classified and incorrectly classified samples, which tend to cluster on the top and bottom half of the image, respectively. On the other hand, variance is a good indicator of OOD and adversarial samples. RED's detection scores of in-distribution samples have low variance because they covary with the training samples. The variance thus represents RED's confidence in its detection score. Samples with large variance indicate that RED is uncertain about its detection score, which can be used as a basis for detecting OOD and adversarial samples.

In order to quantify the potential of RED in detecting OOD and adversarial samples, the variance of detection scores $\mathrm{var}(\hat{r}_*)$ (RED-variance) was used as the detection metric, and detection performance compared with MCP baseline and stardard RED (RED-mean) in all 125 UCI datasets (10 independent runs each). The performance in detecting OOD samples was measured by AP-OOD and AUPR-OOD, which treat OOD samples as the positive class. Similarly, AP-Adversarial and AUPR-Adversarial were used as measures in detecting adversarial samples. The RED training pipeline was exactly the same as in Section~\ref{subsec:exp_UCI}. A summary of the experimental results is shown in Table~\ref{tab:pairwise_comparison_OOD+adv} (see Appendix~\ref{subsec:setup_OOD_adv} for setup details, and Appendix~\ref{subsec:detailed_results_OOD} for detailed results). Intriguingly, RED-variance performs well in both OOD and adversarial sample detection even though it was not trained on any OOD/adversarial samples. In contrast, the original MCP baseline performs significantly worse in both scenarios. The original NN classifier always returns highest class probabilities on deceptive adversarial samples; as a result, MCP makes a purely random guess, resulting in a consistent AP-Adversarial/AUPR-Adversarial of 50\%/25\%. In addition, the comparison between RED-variance and RED-mean verifies that the variance $\mathrm{var}(\hat{r}_*)$ is a more discriminative metric than mean $\hat{c}_*+\bar{\hat{r}}_*$ in detecting OOD and adversarial samples.

The scalability of RED-variance was evaluated in a more complex OOD detection task: Images from the SVHN dataset \citep{Netzer2011} were treated as OOD samples for VGG16 classifiers trained on CIFAR-10 dataset. The same RED and VGG16 models as in Section~\ref{subsec:VGG16_CIFAR10} were used without retraining. Experimental results in Table~\ref{tab:results_CIFAR10vsSVHN} show that RED-variance consistently outperforms the MCP baseline.

Thus, the empirical study in this subsection shows that RED provides a promising foundation not just for detecting misclassifications, but for distinguishing them from other error types as well. This is a new dimension in reliability and interpretability in machine learning systems. RED can therefore serve as a step to make deployments of such systems safer in the future.

\section{Discussion and Future Work}
\label{sec:discussion_future_work}
One interesting observation from the experiments is that RED almost never performs worse than the MCP baseline. This result suggests that there is almost no risk in applying RED on top of an existing NN classifier. Since RED is based on a GP model, the estimated residual $\bar{\hat{r}}_*$ is close to zero if the predicted sample is far from the distribution of the original training samples, resulting in a negligible change to the original MCP. In other words, RED does not make random changes to original MCP if it is very uncertain about the predicted sample, and this uncertainty is explicitly represented in the variance of the estimated detection score. Moreover, the distance-awareness ability of GP also substantially improves the quality of uncertainty estimation \citep{Liu2020}. These properties make RED a particularly reliable technique for error detection.

Another inspiring observation is that the variance is also helpful in detecting OOD and adversarial samples. This result follows from the design of the GP model. Since mRIO in RED has an input kernel and a multi-output kernel, lower estimated variance requires that the predicted sample is close to training samples in both the input feature space and the classifier output space. This requirement is difficult for OOD and adversarial attacks to achieve, providing a basis for detecting them.

In a real-world deployment, it is necessary to define a threshold for triggering error warning based on RED's detection scores. A practical way is to use a validation dataset to determine how the precision/recall tradeoff changes over different thresholds. The user can then select a threshold based on their preference.

The main limitation of RED is that it is not applicable to classifiers that have 100\% prediction accuracy in both training and validation datasets, as might happen in some cases of overfitting to small datasets. In this case, there are no misclassification samples for RED to learn. In practice, this situation is easy to identify by directly looking at the training and validation performance of the original NN classifier. One potential solution is to reserve a portion of data that include at least one misclassification sample for training RED.


The most compelling direction of future work is to extend the capability of RED in distinguishing errors further. Instead of using a single dimensional score for error detection, it is possible to use mean and variance simultaneously, leading to a two dimensional detection space. Further separation between OOD and adversarial samples may be possible by adding one more dimension, such as the ratio between input kernel output and multi-output kernel output. Also, instead of using a hard target detection score (i.e.\ either 0 or 1), it may be possible to define a soft target score that is more informative. Further, RED may be used on top of other existing detection metrics, such as the Trust Score, which may lead to a further improvement in detection performance.

\section{Conclusion}
\label{sec:conclusion}
This paper introduced a new framework, RED, for error detection in neural network classifiers that can produce a more reliable detection score than previous methods. RED is able to not only provide a detection score, but also report the uncertainty of that score. Experimental results show that RED's scores consistently outperform state-of-the-art methods in separating the misclassified samples from correctly classified samples. Further empirical studies also demonstrate that the approach is applicable to various types of base classifiers, scales up to large deep learning architectures, and can form a basis for detecting OOD and adversarial samples as well. It is therefore a promising foundation for improving robustness of neural network classifiers.

\section*{Acknowledgements}
We thank Garrett Bingham for sharing source code and providing helpful suggestions in training WRN.
{\small
\bibliography{ref}
}
\newpage
\appendix
\section{Appendix}
\label{sec:appendix}
\subsection{Experimental Setup for Section~\ref{subsec:exp_UCI} and Section~\ref{subsec:genbase}}
\label{subsec:setup_UCI125}
\paragraph{General Information}
10 independent runs are performed for each dataset. During each run, the dataset is randomly split into a training set ($80\%$) and a testing set ($20\%$), then a fully connected feed-forward NN classifier with 2 hidden layers, each with 64 hidden neurons, are trained on the training set. The activation function is ReLU for all the hidden layers. The maximum number of epochs for training is 1000. $20\%$ of the training set is used as validation set, and the split is random at each independent run. An early stop is triggered if the loss on validation set has not be improved for 10 epochs. The optimizer is Adam with learning rate 0.001, $\beta_1=0.9$, and $\beta_2=0.999$. The loss function is cross entropy loss. During each independent run, the same random dataset split and trained base NN classifier is used for evaluating all algorithms. Results on some datasets are not included in the summary tables if the base classifier does not make any misclassifications, or the number of samples in one particular class is too small for Trust Score to calculate neighborhood distance, or a numerical instability issue happens during the training of the BLR-residual. All source codes for reproducing the experimental results can be found in \href{https://github.com/cognizant-ai-labs/red-paper}{https://github.com/cognizant-ai-labs/red-paper}. All the experiments in this paper are running on a
machine with 20 Intel(R) Xeon(R) Gold 5215 CPU @ 2.50GHz, 128GB memory, and a GTX 2080. 
\paragraph{Dataset Description}
In total, 125 UCI datasets are used in the experiments, among which 121 are from \citet{Klambauer17}, and 4 are recent datasets released in \citet{Dua2017}. All features in all datasets are normalized to have mean 0 and standard deviation 1. Full details regarding the number of samples N, number of features M, and number of classes K for each dataset are shown in Table~\ref{app:table_UCI125_AP-Error}.
\paragraph{Parametric Setup for Algorithms}
\begin{itemize}
	\item RED: SVGP \citep{Hensman2013,Hensman2015} is used as an approximator to original GP. The number of inducing points is 50. RBF kernel is used for both input and multi-output kernel. Automatic Relevance Determination (ARD) feature is turned on. The signal variances and length scales of all the kernels plus the noise variance are the trainable hyperparameters. The optimizer is L-BFGS-B with default parameters as in Scipy.optimize documentation (\href{https://docs.scipy.org/doc/scipy/reference/optimize.minimize-lbfgsb.html}{https://docs.scipy.org/doc/scipy/reference/optimize.minimize-lbfgsb.html}), and the maximum number of iterations is set as 1000. The optimization process runs until the L-BFGS-B optimizer decides to stop. To overcome the sensitivity of GP optimization to initialization of the hyperparameters \citep{Ulapane20}, 20 random initialization of the hyperparameters are tried for each independent run. For each random initialization, the signal variances are generated from a uniform distribution within interval $[0,1]$, and the lengthscales are generated from a uniform distribution within interval $[0,10]$. For 10 initializations, the hyperparameters of input kernel are first optimized while the multi-output kernel is temporarily turned off, then after the optimizer stops, the multi-output kernel is turned on, and both the two kernels are optimized simultaneously. For the other 10 initializations, both kernels are optimized simultaneouly from the start. The average performance of the 3 best optimized model in terms of corresponding metrics are used as the final performance of RED on each independent run. During our preliminary investigation, several statistic metrics on training set is effective in picking the true best-performing model out of these 20 trials, e.g., the gap between average estimated detection scores of correctly classified training samples and incorrectly classified training samples, the scale of optimized noise variance of SVGP model, the ratio between sum of signal variances and noise variance after optimization, etc. Since improving initialization and optimization of GP hyperparameters is out of the scope of this work, we simply use average performance of the best 3 models (top $15\%$) in comparison.
	\item MCP baseline: The maximum class probability of softmax outputs of the base NN classifier is used as the detection score of MCP baseline. The setup of the base NN classifier is provided above.
	\item Trust Score: k=10, $\alpha$=0, without filtering. This is the same as the default setup in \href{https://github.com/google/TrustScore}{https://github.com/google/TrustScore}.
	\item ConfidNet: During training, the input to ConfidNet is the raw feature, and the target is the class probability of the ground-truth class returned by base NN classifier. The architecture of ConfidNet is a fully connected feed-forward NN regressor with 2 hidden layers, each with 64 hidden neurons. The activation function is ReLU for all the hidden layers. The maximum number of epochs for training is 1000. An early stop is triggered if the loss on validation data has not be improved for 10 epochs. The optimizer is RMSprop with learning rate 0.001, and the loss function is mean squared error (MSE).
	\item Introspection-Net: During training, the input to Introspection-Net is the logit outputs of base NN classifier, and the target is 1 for correctly classified sample, and 0 for incorrectly classified sample. The architecture of ConfidNet is a fully connected feed-forward NN regressor with 2 hidden layers, each with 64 hidden neurons. The activation function is ReLU for all the hidden layers. The maximum number of epochs for training is 1000. An early stop is triggered if the loss on validation data has not be improved for 10 epochs. The optimizer is RMSprop with learning rate 0.001, and the loss function is mean squared error (MSE).
	\item Entropy: The entropy of softmax outputs of the base NN classifier is used as the detection score of Entropy. The setup of the base NN classifier is provided above.
	\item DNGO: A Bayesian linear regression layer similar to that of \citet{Snoek2015} is added after the logits layer of the original NN classifier to predict whether an original prediction is correct or not (1 for correct and 0 for incorrect). Default parametric setup, as in \href{https://github.com/automl/pybnn/blob/master/pybnn/dngo.py}{https://github.com/automl/pybnn/blob/master/pybnn/dngo.py}, is used.
	\item SVGP: The original SVGP without output kernel is used to predict directly whether a prediction made by the base NN classifier is correct or not (1 for correct and 0 for incorrect). All other parameters are identical to those in RED.
	\item BNN MCP: The standard dense layers in the base NN classifier described in RED setup above is replaced with Flipout layers \citep{wen2018}. All other parameters are identical with those in RED. The maximum class probability averaging over 100 test-time samplings is used as the detection score for error detection.
	\item BNN Entropy: The same setup as with BNN MCP, except now the entropy of softmax outputs averaging over 100 test-time samplings is used as the detection score for error detection.
	\item MC-Dropout MCP: A dropout layer with dropout rate of 0.5 is added after each dense layer of the base NN classifier described in the RED setup. All other parameters are identical with those in RED. The maximum class probability averaging over 100 test-time Monte-Carlo samplings is used as the detection score for error detection.
	\item MC-Dropout Entropy: The same setup as with MC-Dropout MCP, except now the entropy of softmax outputs is averaged over 100 test-time Monte-Carlo samplings and used as detection score for error detection.
	\item BLR-residual: The GP model in original RED is replaced by a Bayesian linear regression (BLR) similar to that of \citet{Snoek2015}. The BLR is trained to predict the $\bar{\hat{r}}_*$ and $\mathrm{var}(\hat{r}_*)$, and the remaining components in the framework are exactly the same as in the original RED. Default parametric setup, as in \href{https://github.com/automl/pybnn/blob/master/pybnn/dngo.py}{https://github.com/automl/pybnn/blob/master/pybnn/dngo.py}, is used for the BLR.
\end{itemize}

\subsection{Experimental Setup for Section~\ref{subsec:VGG16_CIFAR10}}
\label{subsec:setup_VGG16_CIFAR10}
\paragraph{Setup of VGG16/VGG19 Training} The standard VGG16/VGG19 architectures \citep{Simonyan15} are used. The training pipeline is based on the default setup described in \href{https://github.com/geifmany/cifar-vgg}{https://github.com/geifmany/cifar-vgg}. For the CIFAR-10/CIFAR-100 datasets \citep{Krizhevsky2009}, 40,000 samples are used as the training set, 10,000 as the validation set, and 10,000 as the testing set.
\paragraph{Setup of WRN Training} The standard WRN architecture \citep{ZagoruykoK2016} with a depth of 10 and widening factor of 4 is used. The training pipline is the same as described in \citep{Bingham2020}, and the number of epochs for training is 150 for each run. For the CIFAR-10 dataset \citep{Krizhevsky2009}, 40,000 samples are used as the training set, 10,000 as the validation set, and 10,000 as the testing set.
\paragraph{Parametric Setup for Algorithms} All algorithms use the logit outputs of the trained VGG16/WRN-10-4/VGG19 model as input features. The maximum class probability of softmax outputs of the trained VGG16/WRN-10-4/VGG19 model is used as the detection score of MCP baseline. The parameters for RED, Trust Score, Entropy, DNGO and SVGP are identical to those in the UCI experiments. For ConfidNet and Introspection-Net, all parameters are the same as in the UCI experiments, except for that the number of hidden neurons for all hidden layers is increased to 128.

\subsection{Experimental Setup for Section~\ref{subsec:OOD_adversarial}}
\label{subsec:setup_OOD_adv}
\paragraph{Basic Setup} 10 independent runs are performed for each dataset. The training procedures and parametric setups of RED and MCP are exactly the same as described in Section~\ref{subsec:setup_UCI125}. The only difference is that during testing phase, OOD and adversarial samples are added into the testing dataset, and RED and MCP are required to detect these samples using corresponding detection scores. The numbers of OOD samples, adversarial samples, and in-distribution testing samples are same. When calculating AP-OOD/AUPR-OOD, adversarial samples are excluded. When calculating AP-adversarial/AUPR-adversarial, OOD samples are excluded.
\paragraph{Steps to Generate Adversarial Data} Data points are first generated manually to have exactly the same input features as the original correctly classified training data; incorrect classifications are then assigned manually with highest confidence (class probability 1.0 for a wrong class, and 0 for remaining classes) to each data point as if these were the outputs of the base classifiers. 
\paragraph{Setup of CIFAR-10 vs. SVHN Experiment} The same RED and VGG16 models as in Section~\ref{subsec:VGG16_CIFAR10} are used without retraining. The cropped version (32-by-32 pixels) of SVHN dataset \citep{Netzer2011} is used. 10,000 samples from SVHN test set are randomly selected to be added into the CIFAR-10 testing set, and RED and MCP are required to detect these SVHN samples using corresponding detection scores.

\subsection{Detailed Results for Section~\ref{subsec:exp_UCI} and Section~\ref{subsec:genbase}}
\label{subsec:detailed_results_UCI125}
This subsection shows all the detailed results for experiments performed in Section~\ref{subsec:exp_UCI} and Section~\ref{subsec:genbase}. The results are averaged over 10 independent runs in terms of AP-Error, AP-Success, AUPR-Error, AUPR-Success, and AUROC. Detailed results for Section~\ref{subsec:exp_UCI} are shown in Table~\ref{app:table_UCI125_AP-Error}, Table~\ref{app:table_UCI125_AP-Success}, Table~\ref{app:table_UCI125_AUPR-Error}, Table~\ref{app:table_UCI125_AUPR-Success}, and Table\ref{app:table_UCI125_AUROC}. Detailed results for Section~\ref{subsec:genbase} are shown in Table~\ref{app:table_UCI125_AP-Error_BNN+dropout}, Table~\ref{app:table_UCI125_AP-Success_BNN+dropout}, Table~\ref{app:table_UCI125_AUPR-Error_BNN+dropout}, Table~\ref{app:table_UCI125_AUPR-Success_BNN+dropout}, and Table\ref{app:table_UCI125_AUROC_BNN+dropout}. Each table is corresponding to one performance metric. The column "N", "M", and "K" denotes the number of samples, number of features, and number of classes for corresponding datasets. To save space, "MCP", "Intro-Net", "MC-D" stands for "MCP Baseline", "Introspection-Net", and "MC-Dropout", respectively. For datasets that the original NN classifier achieves $100\%$ accuracy, the entries are marked as "NA". For dataset splits that the number of samples in one particular class is too small for Trust Score to calculate neighborhood distance, the entries are marked as "NA".

\onecolumn
{\scriptsize
	\setlength{\tabcolsep}{3.8pt}
\begin{longtable}{l c c c c c c c c c c c}
		\caption{Comparison between RED and Counterparts Using AP-Error} \label{app:table_UCI125_AP-Error}\\
		\toprule
		Dataset & N & M & K &RED & MCP & Trust Score & ConfidNet & Intro-Net & Entropy & DNGO & SVGP\\ \midrule
		\endfirsthead
		\toprule
		Dataset & N & M & K &RED & MCP & Trust Score & ConfidNet & Intro-Net & Entropy & DNGO & SVGP \\ \midrule
		\endhead
		\\\emph{Continued on next page.} & & & & & & & &\\
		\endfoot
		\\\emph{Continued from previous page.} & & & & & & & &\\
		\endlastfoot
		abalone&4177&8&3&0.546&0.539&0.480&0.518&0.528&0.529&0.496&0.482\\
		acute-inflammation&120&6&2&NA&NA&NA&NA&NA&NA&NA&NA\\
		acute-nephritis&120&6&2&NA&NA&NA&NA&NA&NA&NA&NA\\
		adult&48842&14&2&0.419&0.411&0.328&0.391&0.412&0.411&0.338&0.358\\
		annealing&898&31&5&0.558&0.392&0.453&0.350&0.399&0.368&0.314&0.324\\
		arrhythmia&452&262&13&0.583&0.580&0.618&0.409&0.413&0.560&0.426&0.475\\
		audiology-std&196&59&18&0.779&0.721&NA&0.428&0.518&0.731&0.590&0.556\\
		balance-scale&625&4&3&0.888&0.816&0.540&0.046&0.155&0.730&0.319&0.128\\
\bottomrule
\end{longtable}}

{\scriptsize
	\setlength{\tabcolsep}{3.8pt}
	\begin{longtable}{l c c c c c c c c c c c}
		\caption{Comparison between RED and Counterparts Using AP-Success} \label{app:table_UCI125_AP-Success}\\
		\toprule
		Dataset & N & M & K &RED & MCP & Trust Score & ConfidNet & Intro-Net & Entropy & DNGO & SVGP\\ \midrule
		\endfirsthead
		\toprule
		Dataset & N & M & K &RED & MCP & Trust Score & ConfidNet & Intro-Net & Entropy & DNGO & SVGP \\ \midrule
		\endhead
		\\\emph{Continued on next page.} & & & & & & & &\\
		\endfoot
		\\\emph{Continued from previous page.} & & & & & & & &\\
		\endlastfoot
		abalone&4177&8&3&0.853&0.850&0.821&0.845&0.844&0.849&0.814&0.840\\
		acute-inflammation&120&6&2&NA&NA&NA&NA&NA&NA&NA&NA\\
		acute-nephritis&120&6&2&NA&NA&NA&NA&NA&NA&NA&NA\\
		adult&48842&14&2&0.968&0.967&0.944&0.964&0.961&0.967&0.927&0.961\\
		annealing&898&31&5&0.975&0.899&0.943&0.909&0.916&0.902&0.898&0.913\\
		arrhythmia&452&262&13&0.817&0.810&0.809&0.654&0.668&0.803&0.673&0.778\\
		audiology-std&196&59&18&0.946&0.926&NA&0.625&0.740&0.927&0.756&0.782\\
		balance-scale&625&4&3&0.999&0.999&0.997&0.972&0.983&0.999&0.985&0.976\\
\bottomrule
\end{longtable}}

{\scriptsize
	\setlength{\tabcolsep}{3.8pt}
	\begin{longtable}{l c c c c c c c c c c c}
		\caption{Comparison between RED and Counterparts Using AUPR-Error} \label{app:table_UCI125_AUPR-Error}\\
		\toprule
		Dataset & N & M & K &RED & MCP & Trust Score & ConfidNet & Intro-Net & Entropy & DNGO & SVGP\\ \midrule
		\endfirsthead
		\toprule
		Dataset & N & M & K &RED & MCP & Trust Score & ConfidNet & Intro-Net & Entropy & DNGO & SVGP \\ \midrule
		\endhead
		\\\emph{Continued on next page.} & & & & & & & &\\
		\endfoot
		\\\emph{Continued from previous page.} & & & & & & & &\\
		\endlastfoot
		abalone&4177&8&3&0.543&0.537&0.478&0.516&0.526&0.527&0.498&0.480\\
		acute-inflammation&120&6&2&NA&NA&NA&NA&NA&NA&NA&NA\\
		acute-nephritis&120&6&2&NA&NA&NA&NA&NA&NA&NA&NA\\
		adult&48842&14&2&0.419&0.410&0.328&0.390&0.412&0.410&0.372&0.357\\
		annealing&898&31&5&0.547&0.375&0.440&0.331&0.385&0.350&0.347&0.352\\
		arrhythmia&452&262&13&0.573&0.570&0.606&0.396&0.399&0.547&0.412&0.465\\
		audiology-std&196&59&18&0.766&0.704&NA&0.401&0.494&0.712&0.564&0.532\\
		balance-scale&625&4&3&0.877&0.794&0.501&0.030&0.124&0.685&0.396&0.085\\
\bottomrule
\end{longtable}}

{\scriptsize
	\setlength{\tabcolsep}{3.8pt}
	\begin{longtable}{l c c c c c c c c c c c}
		\caption{Comparison between RED and Counterparts Using AUPR-Success} \label{app:table_UCI125_AUPR-Success}\\
		\toprule
		Dataset & N & M & K &RED & MCP & Trust Score & ConfidNet & Intro-Net & Entropy & DNGO & SVGP\\ \midrule
		\endfirsthead
		\toprule
		Dataset & N & M & K &RED & MCP & Trust Score & ConfidNet & Intro-Net & Entropy & DNGO & SVGP \\ \midrule
		\endhead
		\\\emph{Continued on next page.} & & & & & & & &\\
		\endfoot
		\\\emph{Continued from previous page.} & & & & & & & &\\
		\endlastfoot
		abalone&4177&8&3&0.852&0.850&0.821&0.844&0.844&0.849&0.806&0.840\\
		acute-inflammation&120&6&2&NA&NA&NA&NA&NA&NA&NA&NA\\
		acute-nephritis&120&6&2&NA&NA&NA&NA&NA&NA&NA&NA\\
		adult&48842&14&2&0.968&0.967&0.944&0.964&0.961&0.967&0.897&0.961\\
		annealing&898&31&5&0.975&0.903&0.942&0.907&0.914&0.899&0.908&0.919\\
		arrhythmia&452&262&13&0.814&0.806&0.806&0.645&0.659&0.799&0.665&0.775\\
		audiology-std&196&59&18&0.945&0.925&NA&0.609&0.725&0.925&0.741&0.769\\
		balance-scale&625&4&3&0.999&0.999&0.997&0.971&0.983&0.999&0.986&0.976\\
\bottomrule
\end{longtable}}

{\scriptsize
	\setlength{\tabcolsep}{3.8pt}
	\begin{longtable}{l c c c c c c c c c c c}
		\caption{Comparison between RED and Counterparts Using AUROC} \label{app:table_UCI125_AUROC}\\
		\toprule
		Dataset & N & M & K &RED & MCP & Trust Score & ConfidNet & Intro-Net & Entropy & DNGO & SVGP\\ \midrule
		\endfirsthead
		\toprule
		Dataset & N & M & K &RED & MCP & Trust Score & ConfidNet & Intro-Net & Entropy & DNGO & SVGP \\ \midrule
		\endhead
		\\\emph{Continued on next page.} & & & & & & & &\\
		\endfoot
		\\\emph{Continued from previous page.} & & & & & & & &\\
		\endlastfoot
		abalone&4177&8&3&0.734&0.730&0.680&0.719&0.726&0.726&0.697&0.700\\
		acute-inflammation&120&6&2&NA&NA&NA&NA&NA&NA&NA&NA\\
		acute-nephritis&120&6&2&NA&NA&NA&NA&NA&NA&NA&NA\\
		adult&48842&14&2&0.841&0.837&0.759&0.828&0.830&0.837&0.751&0.814\\
		annealing&898&31&5&0.865&0.699&0.776&0.724&0.723&0.694&0.619&0.675\\
		arrhythmia&452&262&13&0.712&0.715&0.727&0.505&0.517&0.703&0.528&0.639\\
		audiology-std&196&59&18&0.883&0.848&NA&0.439&0.605&0.852&0.666&0.672\\
		balance-scale&625&4&3&0.986&0.987&0.909&0.510&0.621&0.985&0.712&0.536\\
\bottomrule
\end{longtable}}

{\scriptsize
	\setlength{\tabcolsep}{3.8pt}
	\begin{longtable}{l c c c c c c c c c}
		\caption{Comparison between RED and Counterparts Using AP-Error} \label{app:table_UCI125_AP-Error_BNN+dropout}\\
		\toprule
		Dataset & N & M & K & RED+BNN & BNN MCP & BNN Entropy & RED+MC-D & MC-D MCP & MC-D Entropy \\ \midrule
		\endfirsthead
		\toprule
		Dataset & N & M & K & RED+BNN & BNN MCP & BNN Entropy & RED+MC-D & MC-D MCP & MC-D Entropy \\ \midrule
		\endhead
		\\\emph{Continued on next page.} & & & & & & & & &\\
		\endfoot
		\\\emph{Continued from previous page.} & & & & & & & & &\\
		\endlastfoot
		abalone&4177&8&3&0.529&0.522&0.513&0.516&0.509&0.503\\
		acute-inflammation&120&6&2&NA&NA&NA&NA&NA&NA\\
		acute-nephritis&120&6&2&NA&NA&NA&NA&NA&NA\\
		adult&48842&14&2&0.415&0.407&0.406&0.416&0.405&0.405\\
		annealing&898&31&5&0.788&0.424&0.338&0.590&0.421&0.375\\
		arrhythmia&452&262&13&0.719&0.614&0.734&0.551&0.543&0.540\\
		audiology-std&196&59&18&0.768&0.684&0.756&0.760&0.659&0.698\\
		balance-scale&625&4&3&0.820&0.532&0.581&0.685&0.548&0.621\\
\bottomrule
\end{longtable}}

{\scriptsize
	\setlength{\tabcolsep}{3.8pt}
	\begin{longtable}{l c c c c c c c c c}
		\caption{Comparison between RED and Counterparts Using AP-Success} \label{app:table_UCI125_AP-Success_BNN+dropout}\\
		\toprule
		Dataset & N & M & K & RED+BNN & BNN MCP & BNN Entropy & RED+MC-D & MC-D MCP & MC-D Entropy \\ \midrule
		\endfirsthead
		\toprule
		Dataset & N & M & K & RED+BNN & BNN MCP & BNN Entropy & RED+MC-D & MC-D MCP & MC-D Entropy \\ \midrule
		\endhead
		\\\emph{Continued on next page.} & & & & & & & & &\\
		\endfoot
		\\\emph{Continued from previous page.} & & & & & & & & &\\
		\endlastfoot
		abalone&4177&8&3&0.855&0.854&0.849&0.854&0.852&0.851\\
		acute-inflammation&120&6&2&NA&NA&NA&NA&NA&NA\\
		acute-nephritis&120&6&2&NA&NA&NA&NA&NA&NA\\
		adult&48842&14&2&0.970&0.969&0.970&0.970&0.969&0.970\\
		annealing&898&31&5&0.983&0.950&0.947&0.981&0.971&0.974\\
		arrhythmia&452&262&13&0.805&0.756&0.811&0.831&0.814&0.827\\
		audiology-std&196&59&18&0.920&0.881&0.939&0.943&0.906&0.927\\
		balance-scale&625&4&3&0.999&0.996&0.997&0.999&0.998&0.999\\
\bottomrule
\end{longtable}}

{\scriptsize
	\setlength{\tabcolsep}{3.8pt}
	\begin{longtable}{l c c c c c c c c c}
		\caption{Comparison between RED and Counterparts Using AUPR-Error} \label{app:table_UCI125_AUPR-Error_BNN+dropout}\\
		\toprule
		Dataset & N & M & K & RED+BNN & BNN MCP & BNN Entropy & RED+MC-D & MC-D MCP & MC-D Entropy \\ \midrule
		\endfirsthead
		\toprule
		Dataset & N & M & K & RED+BNN & BNN MCP & BNN Entropy & RED+MC-D & MC-D MCP & MC-D Entropy \\ \midrule
		\endhead
		\\\emph{Continued on next page.} & & & & & & & & &\\
		\endfoot
		\\\emph{Continued from previous page.} & & & & & & & & &\\
		\endlastfoot
		abalone&4177&8&3&0.526&0.519&0.510&0.513&0.506&0.501\\
		acute-inflammation&120&6&2&NA&NA&NA&NA&NA&NA\\
		acute-nephritis&120&6&2&NA&NA&NA&NA&NA&NA\\
		adult&48842&14&2&0.414&0.406&0.405&0.415&0.405&0.404\\
		annealing&898&31&5&0.785&0.414&0.327&0.580&0.401&0.353\\
		arrhythmia&452&262&13&0.713&0.603&0.726&0.539&0.531&0.529\\
		audiology-std&196&59&18&0.758&0.683&0.744&0.748&0.637&0.674\\
		balance-scale&625&4&3&0.790&0.474&0.518&0.656&0.519&0.580\\
\bottomrule
\end{longtable}}

{\scriptsize
	\setlength{\tabcolsep}{3.8pt}
	\begin{longtable}{l c c c c c c c c c}
		\caption{Comparison between RED and Counterparts Using AUPR-Success} \label{app:table_UCI125_AUPR-Success_BNN+dropout}\\
		\toprule
		Dataset & N & M & K & RED+BNN & BNN MCP & BNN Entropy & RED+MC-D & MC-D MCP & MC-D Entropy \\ \midrule
		\endfirsthead
		\toprule
		Dataset & N & M & K & RED+BNN & BNN MCP & BNN Entropy & RED+MC-D & MC-D MCP & MC-D Entropy \\ \midrule
		\endhead
		\\\emph{Continued on next page.} & & & & & & & & &\\
		\endfoot
		\\\emph{Continued from previous page.} & & & & & & & & &\\
		\endlastfoot
		abalone&4177&8&3&0.855&0.853&0.849&0.853&0.852&0.851\\
		acute-inflammation&120&6&2&NA&NA&NA&NA&NA&NA\\
		acute-nephritis&120&6&2&NA&NA&NA&NA&NA&NA\\
		adult&48842&14&2&0.970&0.969&0.970&0.970&0.969&0.970\\
		annealing&898&31&5&0.983&0.951&0.947&0.981&0.972&0.974\\
		arrhythmia&452&262&13&0.801&0.750&0.806&0.827&0.809&0.823\\
		audiology-std&196&59&18&0.918&0.912&0.938&0.942&0.902&0.925\\
		balance-scale&625&4&3&0.999&0.996&0.997&0.999&0.998&0.999\\
\bottomrule
\end{longtable}}

{\scriptsize
	\setlength{\tabcolsep}{3.8pt}
	\begin{longtable}{l c c c c c c c c c}
		\caption{Comparison between RED and Counterparts Using AUROC} \label{app:table_UCI125_AUROC_BNN+dropout}\\
		\toprule
		Dataset & N & M & K & RED+BNN & BNN MCP & BNN Entropy & RED+MC-D & MC-D MCP & MC-D Entropy \\ \midrule
		\endfirsthead
		\toprule
		Dataset & N & M & K & RED+BNN & BNN MCP & BNN Entropy & RED+MC-D & MC-D MCP & MC-D Entropy \\ \midrule
		\endhead
		\\\emph{Continued on next page.} & & & & & & & & &\\
		\endfoot
		\\\emph{Continued from previous page.} & & & & & & & & &\\
		\endlastfoot
		abalone&4177&8&3&0.731&0.727&0.718&0.725&0.723&0.716\\
		acute-inflammation&120&6&2&NA&NA&NA&NA&NA&NA\\
		acute-nephritis&120&6&2&NA&NA&NA&NA&NA&NA\\
		adult&48842&14&2&0.845&0.842&0.843&0.846&0.841&0.841\\
		annealing&898&31&5&0.928&0.798&0.764&0.883&0.850&0.836\\
		arrhythmia&452&262&13&0.766&0.703&0.784&0.709&0.698&0.700\\
		audiology-std&196&59&18&0.850&0.814&0.865&0.874&0.814&0.851\\
		balance-scale&625&4&3&0.982&0.946&0.965&0.977&0.955&0.968\\
\bottomrule
\end{longtable}}

\subsection{Detailed Results for Section~\ref{subsec:OOD_adversarial}}
\label{subsec:detailed_results_OOD}
This subsection shows all the detailed results for experiments performed in Section~\ref{subsec:OOD_adversarial}. The results are averaged over 10 independent runs. Table~\ref{app:table_UCI125_OOD} shows the results for OOD detection, and Table~\ref{app:table_UCI125_adversarial} shows the results for detecting adversarial samples. The original NN classifier always returns highest class probabilities on deceptive adversarial samples; as a result MCP makes a purely random guess, resulting in a consistent AP-Adversarial/AUPR-Adversarial of 50\%/25\%.
\onecolumn
{\scriptsize
	\begin{longtable}{l c c}
		\caption{Comparison between RED and Counterparts on OOD detection (mean$\pm$std over 10 runs)} \label{app:table_UCI125_OOD}\\
		\toprule
		\multirow{2}{*}{Dataset} & AP-OOD(\%) & AUPR-OOD(\%)\\
		{}	& RED-variance / MCP baseline / RED-mean & RED-variance / MCP baseline / RED-mean\\
		\midrule
		\endfirsthead
		\toprule
		\multirow{2}{*}{Dataset} & AP-OOD(\%) & AUPR-OOD(\%)\\
		{}	& RED-variance / MCP baseline / RED-mean & RED-variance / MCP baseline / RED-mean\\
		\midrule
		\endhead
		\\\emph{Continued on next page.} & & \\
		\endfoot
		\\\emph{Continued from previous page.} & & \\
		\endlastfoot
		abalone&83.93$\pm$9.58 / 41.95$\pm$1.64 / 42.29$\pm$1.66&83.90$\pm$9.60 / 41.86$\pm$1.64 / 42.20$\pm$1.66\\
		acute-inflammation&89.83$\pm$3.31 / 88.46$\pm$4.22 / 89.21$\pm$4.12&89.66$\pm$3.37 / 88.38$\pm$4.38 / 89.06$\pm$4.18\\
		acute-nephritis&91.88$\pm$4.80 / 91.76$\pm$4.58 / 93.27$\pm$3.16&91.75$\pm$4.88 / 91.87$\pm$4.36 / 93.17$\pm$3.21\\
		adult&93.59$\pm$5.25 / 49.78$\pm$0.79 / 50.47$\pm$1.02&93.59$\pm$5.26 / 49.76$\pm$0.79 / 50.46$\pm$1.02\\
		annealing&85.71$\pm$5.05 / 59.85$\pm$3.54 / 75.81$\pm$4.78&85.57$\pm$5.12 / 59.64$\pm$3.59 / 75.67$\pm$4.82\\
		arrhythmia&62.85$\pm$3.90 / 64.34$\pm$3.96 / 65.85$\pm$4.15&63.09$\pm$4.67 / 63.91$\pm$4.00 / 65.47$\pm$4.20\\
		audiology-std&85.71$\pm$11.02 / 68.11$\pm$7.19 / 73.85$\pm$7.47&85.29$\pm$11.34 / 67.14$\pm$7.61 / 73.31$\pm$7.59\\
		balance-scale&57.87$\pm$5.53 / 52.00$\pm$1.93 / 56.20$\pm$2.52&57.42$\pm$5.60 / 51.55$\pm$2.91 / 55.69$\pm$2.61\\
		balloons&100.00$\pm$0.00 / 51.85$\pm$14.39 / 59.78$\pm$13.93&100.00$\pm$0.00 / 42.69$\pm$15.96 / 51.13$\pm$16.39\\
		bank&88.56$\pm$5.25 / 54.65$\pm$1.87 / 58.24$\pm$2.34&88.51$\pm$5.25 / 54.55$\pm$1.86 / 58.15$\pm$2.34\\
		bioconcentration&90.43$\pm$7.71 / 45.54$\pm$1.15 / 47.48$\pm$1.49&90.42$\pm$7.71 / 45.45$\pm$1.16 / 47.47$\pm$1.49\\
		blood&65.86$\pm$4.32 / 47.16$\pm$3.08 / 47.18$\pm$3.07&65.43$\pm$4.30 / 46.72$\pm$3.10 / 46.75$\pm$3.10\\
		breast-cancer&90.46$\pm$12.06 / 54.66$\pm$4.80 / 60.41$\pm$5.61&90.17$\pm$12.42 / 53.62$\pm$4.81 / 59.66$\pm$5.87\\
		breast-cancer-wisc&87.64$\pm$4.11 / 83.56$\pm$3.54 / 84.26$\pm$3.64&87.51$\pm$4.16 / 83.40$\pm$3.64 / 84.08$\pm$3.69\\
		breast-cancer-wisc-diag&80.60$\pm$6.22 / 78.45$\pm$3.89 / 78.53$\pm$4.61&80.32$\pm$6.34 / 78.28$\pm$3.84 / 78.17$\pm$4.73\\
		breast-cancer-wisc-prog&67.02$\pm$4.35 / 50.80$\pm$4.60 / 52.56$\pm$5.86&65.91$\pm$4.50 / 49.49$\pm$4.62 / 51.29$\pm$5.91\\
		breast-tissue&81.55$\pm$7.55 / 55.17$\pm$8.15 / 59.37$\pm$9.42&80.59$\pm$7.97 / 53.62$\pm$8.67 / 57.81$\pm$10.01\\
		car&83.22$\pm$15.61 / 58.61$\pm$2.58 / 61.13$\pm$3.04&83.14$\pm$15.70 / 59.25$\pm$2.66 / 60.93$\pm$2.98\\
		cardiotocography-10clases&78.06$\pm$8.07 / 56.56$\pm$1.91 / 59.41$\pm$3.48&77.96$\pm$8.09 / 56.43$\pm$1.92 / 59.29$\pm$3.49\\
		cardiotocography-3clases&82.64$\pm$8.53 / 50.59$\pm$1.22 / 55.83$\pm$2.99&82.49$\pm$8.53 / 50.46$\pm$1.27 / 55.67$\pm$2.99\\
		chess-krvk&61.45$\pm$5.02 / 50.09$\pm$0.63 / 50.54$\pm$0.81&61.44$\pm$5.02 / 50.06$\pm$0.64 / 50.52$\pm$0.81\\
		chess-krvkp&97.56$\pm$1.37 / 62.89$\pm$1.50 / 74.42$\pm$4.18&97.58$\pm$1.37 / 63.82$\pm$1.72 / 74.27$\pm$4.20\\
		climate&93.89$\pm$1.56 / 45.65$\pm$0.73 / 48.58$\pm$0.78&93.89$\pm$1.56 / 45.56$\pm$0.75 / 48.57$\pm$0.78\\
		congressional-voting&90.47$\pm$3.02 / 41.82$\pm$2.95 / 41.83$\pm$2.94&90.12$\pm$3.13 / 41.45$\pm$2.96 / 41.45$\pm$2.95\\
		conn-bench-sonar-mines-rocks&61.83$\pm$5.34 / 67.13$\pm$5.79 / 67.28$\pm$5.58&60.93$\pm$5.52 / 66.40$\pm$5.94 / 66.54$\pm$5.75\\
		conn-bench-vowel-deterding&75.03$\pm$6.96 / 81.25$\pm$4.41 / 81.93$\pm$4.10&74.89$\pm$7.00 / 81.20$\pm$4.42 / 81.81$\pm$4.17\\
		connect-4&89.32$\pm$10.57 / 45.64$\pm$0.88 / 47.11$\pm$1.16&89.31$\pm$10.57 / 45.54$\pm$0.89 / 47.10$\pm$1.16\\
		contrac&87.02$\pm$9.48 / 45.54$\pm$1.77 / 46.15$\pm$1.64&86.97$\pm$9.51 / 45.25$\pm$1.78 / 45.90$\pm$1.64\\
		credit-approval&96.62$\pm$2.58 / 62.77$\pm$4.52 / 64.55$\pm$3.33&96.57$\pm$2.63 / 62.37$\pm$4.57 / 64.23$\pm$3.37\\
		cylinder-bands&91.81$\pm$7.64 / 54.83$\pm$5.47 / 61.57$\pm$7.26&91.68$\pm$7.73 / 54.28$\pm$5.61 / 61.07$\pm$7.38\\
		dermatology&94.29$\pm$3.33 / 95.15$\pm$1.22 / 95.64$\pm$1.50&94.19$\pm$3.38 / 95.11$\pm$1.24 / 95.60$\pm$1.52\\
		echocardiogram&88.13$\pm$5.07 / 61.17$\pm$7.11 / 63.57$\pm$7.34&87.58$\pm$5.30 / 59.68$\pm$7.56 / 62.32$\pm$7.78\\
		ecoli&85.73$\pm$8.27 / 72.82$\pm$4.98 / 73.21$\pm$4.81&85.46$\pm$8.41 / 72.53$\pm$5.04 / 72.93$\pm$4.86\\
		energy-y1&89.24$\pm$11.65 / 61.29$\pm$3.44 / 66.68$\pm$3.52&89.14$\pm$11.76 / 61.29$\pm$3.45 / 66.41$\pm$3.57\\
		energy-y2&91.05$\pm$9.09 / 61.38$\pm$3.29 / 64.39$\pm$3.06&90.98$\pm$9.16 / 61.32$\pm$3.27 / 64.10$\pm$3.08\\
		fertility&87.69$\pm$14.21 / 54.77$\pm$9.28 / 59.31$\pm$11.36&87.10$\pm$14.89 / 52.36$\pm$9.64 / 57.18$\pm$11.92\\
		flags&96.71$\pm$2.78 / 58.26$\pm$5.04 / 71.85$\pm$7.41&96.60$\pm$2.94 / 57.07$\pm$5.21 / 71.31$\pm$7.55\\
		glass&75.75$\pm$6.71 / 67.01$\pm$5.58 / 70.37$\pm$6.85&75.00$\pm$7.03 / 66.28$\pm$5.87 / 69.79$\pm$7.08\\
		haberman-survival&69.84$\pm$8.57 / 58.38$\pm$5.19 / 58.82$\pm$5.59&69.17$\pm$8.72 / 57.63$\pm$5.29 / 58.09$\pm$5.70\\
		hayes-roth&72.95$\pm$10.37 / 53.92$\pm$7.09 / 55.03$\pm$8.23&72.20$\pm$10.67 / 52.53$\pm$7.36 / 53.51$\pm$8.53\\
		heart-cleveland&92.10$\pm$7.17 / 54.49$\pm$3.41 / 59.15$\pm$4.61&91.95$\pm$7.33 / 53.69$\pm$3.35 / 58.40$\pm$4.63\\
		heart-hungarian&95.37$\pm$4.10 / 65.82$\pm$3.01 / 68.73$\pm$4.71&95.26$\pm$4.23 / 65.16$\pm$3.15 / 68.10$\pm$4.88\\
		heart-switzerland&92.95$\pm$6.37 / 51.05$\pm$5.08 / 62.39$\pm$10.45&92.65$\pm$6.71 / 49.26$\pm$5.37 / 61.03$\pm$10.95\\
		heart-va&91.03$\pm$12.24 / 52.79$\pm$4.81 / 65.28$\pm$9.50&90.71$\pm$12.72 / 51.57$\pm$5.10 / 64.37$\pm$9.96\\
		hepatitis&95.96$\pm$4.06 / 53.59$\pm$4.66 / 65.30$\pm$7.40&95.55$\pm$4.65 / 51.71$\pm$4.58 / 63.76$\pm$7.72\\
		hill-valley&96.53$\pm$2.94 / 32.79$\pm$1.43 / 41.89$\pm$12.72&96.52$\pm$2.94 / 32.54$\pm$1.43 / 41.73$\pm$12.69\\
		horse-colic&86.76$\pm$7.75 / 56.60$\pm$2.83 / 58.82$\pm$4.01&86.58$\pm$7.88 / 55.83$\pm$2.95 / 58.09$\pm$4.08\\
		ilpd-indian-liver&83.55$\pm$7.05 / 39.35$\pm$1.86 / 43.32$\pm$4.26&83.34$\pm$7.16 / 38.87$\pm$1.77 / 42.93$\pm$4.32\\
		image-segmentation&89.77$\pm$7.00 / 73.62$\pm$1.58 / 78.12$\pm$2.42&89.76$\pm$7.03 / 73.64$\pm$1.64 / 78.03$\pm$2.45\\
		ionosphere&80.48$\pm$6.49 / 68.67$\pm$5.58 / 68.97$\pm$4.67&80.08$\pm$6.67 / 68.14$\pm$5.75 / 68.41$\pm$4.76\\
		iris&82.61$\pm$8.19 / 67.10$\pm$4.57 / 69.73$\pm$5.38&82.18$\pm$8.38 / 65.77$\pm$4.98 / 68.59$\pm$5.94\\
		led-display&93.26$\pm$7.93 / 77.49$\pm$1.40 / 77.51$\pm$1.37&93.23$\pm$7.98 / 77.38$\pm$1.42 / 77.41$\pm$1.39\\
		lenses&95.67$\pm$8.67 / 67.25$\pm$4.40 / 67.69$\pm$6.94&95.36$\pm$9.28 / 64.07$\pm$5.03 / 63.42$\pm$9.03\\
		letter&68.99$\pm$5.85 / 70.26$\pm$0.98 / 69.43$\pm$1.42&68.97$\pm$5.85 / 70.43$\pm$0.99 / 69.42$\pm$1.42\\
		libras&78.47$\pm$5.72 / 90.96$\pm$3.67 / 92.29$\pm$2.56&78.24$\pm$5.81 / 90.85$\pm$3.75 / 92.23$\pm$2.59\\
		low-res-spect&68.18$\pm$8.57 / 70.64$\pm$6.42 / 72.22$\pm$5.88&67.82$\pm$8.64 / 70.24$\pm$6.59 / 71.87$\pm$6.05\\
		lung-cancer&91.40$\pm$8.01 / 64.06$\pm$15.83 / 75.22$\pm$13.07&90.27$\pm$9.46 / 60.49$\pm$17.42 / 71.83$\pm$15.32\\
		lymphography&97.04$\pm$5.84 / 64.82$\pm$6.65 / 71.75$\pm$8.23&96.77$\pm$6.37 / 63.73$\pm$7.15 / 70.95$\pm$8.62\\
		magic&74.47$\pm$6.50 / 46.07$\pm$2.80 / 47.70$\pm$2.00&74.47$\pm$6.50 / 45.86$\pm$2.85 / 47.67$\pm$2.00\\
		mammographic&88.12$\pm$9.06 / 51.59$\pm$3.07 / 52.94$\pm$3.56&88.07$\pm$9.10 / 51.24$\pm$3.07 / 52.60$\pm$3.54\\
		messidor&85.66$\pm$11.46 / 44.90$\pm$0.89 / 47.34$\pm$2.06&85.66$\pm$11.46 / 44.82$\pm$0.88 / 47.33$\pm$2.06\\
		miniboone&99.36$\pm$0.58 / 44.37$\pm$0.60 / 41.75$\pm$4.89&99.35$\pm$0.58 / 28.46$\pm$0.45 / 40.68$\pm$5.63\\
		molec-biol-promoter&72.28$\pm$8.19 / 56.94$\pm$11.01 / 58.61$\pm$9.33&71.60$\pm$8.61 / 55.07$\pm$11.47 / 56.82$\pm$9.76\\
		molec-biol-splice&53.15$\pm$4.22 / 52.13$\pm$1.30 / 52.14$\pm$1.28&53.05$\pm$4.24 / 52.01$\pm$1.32 / 52.00$\pm$1.29\\
		monks-1&94.62$\pm$5.15 / 62.79$\pm$2.56 / 71.85$\pm$8.26&94.58$\pm$5.19 / 62.50$\pm$2.61 / 71.52$\pm$8.30\\
		monks-2&88.52$\pm$11.62 / 48.82$\pm$5.15 / 52.59$\pm$7.84&88.45$\pm$11.72 / 48.28$\pm$5.21 / 52.06$\pm$7.96\\
		monks-3&95.36$\pm$8.70 / 51.76$\pm$2.92 / 60.83$\pm$6.50&95.31$\pm$8.81 / 51.13$\pm$3.03 / 60.31$\pm$6.49\\
		mushroom&97.05$\pm$2.23 / 72.89$\pm$1.39 / 85.14$\pm$3.69&97.05$\pm$2.23 / 86.39$\pm$0.70 / 85.14$\pm$3.69\\
		musk-1&60.98$\pm$5.99 / 67.07$\pm$3.37 / 67.20$\pm$3.40&60.47$\pm$5.96 / 66.49$\pm$3.43 / 66.62$\pm$3.47\\
		musk-2&60.39$\pm$3.15 / 58.98$\pm$2.09 / 62.37$\pm$2.38&59.31$\pm$3.09 / 61.01$\pm$2.53 / 62.34$\pm$2.36\\
		nursery&72.44$\pm$9.00 / 60.49$\pm$1.05 / 63.44$\pm$1.58&72.43$\pm$9.00 / 65.00$\pm$1.31 / 63.42$\pm$1.59\\
		oocytes\_merluccius\_nucleus\_4d&86.09$\pm$9.85 / 40.20$\pm$1.85 / 41.84$\pm$1.82&85.97$\pm$9.94 / 39.78$\pm$1.88 / 41.48$\pm$1.80\\
		oocytes\_merluccius\_states\_2f&78.90$\pm$11.01 / 55.58$\pm$3.27 / 57.23$\pm$2.70&78.74$\pm$11.10 / 55.64$\pm$3.17 / 56.95$\pm$2.74\\
		oocytes\_trisopterus\_nucleus\_2f&76.41$\pm$12.52 / 39.96$\pm$1.62 / 41.31$\pm$1.73&76.23$\pm$12.62 / 39.58$\pm$1.58 / 40.96$\pm$1.74\\
		oocytes\_trisopterus\_states\_5b&80.92$\pm$7.10 / 54.69$\pm$4.05 / 56.80$\pm$3.10&80.69$\pm$7.15 / 54.33$\pm$4.12 / 56.48$\pm$3.17\\
		optical&86.68$\pm$5.17 / 92.85$\pm$0.36 / 93.04$\pm$0.35&86.58$\pm$5.31 / 92.86$\pm$0.35 / 93.03$\pm$0.35\\
		ozone&58.40$\pm$7.02 / 45.38$\pm$1.56 / 48.26$\pm$2.79&58.28$\pm$7.01 / 45.31$\pm$1.77 / 48.12$\pm$2.79\\
		page-blocks&90.50$\pm$3.11 / 64.22$\pm$1.79 / 71.06$\pm$4.01&90.45$\pm$3.12 / 63.70$\pm$1.89 / 71.01$\pm$4.01\\
		parkinsons&79.34$\pm$9.66 / 53.44$\pm$4.38 / 59.03$\pm$3.96&78.73$\pm$10.00 / 51.88$\pm$4.71 / 57.88$\pm$4.23\\
		pendigits&79.27$\pm$9.45 / 90.01$\pm$0.82 / 89.96$\pm$0.77&79.26$\pm$9.45 / 90.14$\pm$0.83 / 89.95$\pm$0.77\\
		phishing&89.05$\pm$10.20 / 45.11$\pm$1.14 / 46.98$\pm$1.56&89.04$\pm$10.20 / 45.01$\pm$1.16 / 46.97$\pm$1.56\\
		pima&63.16$\pm$6.00 / 57.61$\pm$1.67 / 57.85$\pm$1.69&62.76$\pm$6.03 / 57.17$\pm$1.71 / 57.39$\pm$1.76\\
		pittsburg-bridges-MATERIAL&88.53$\pm$16.82 / 59.72$\pm$4.83 / 66.64$\pm$10.14&88.06$\pm$17.48 / 57.99$\pm$5.13 / 65.28$\pm$10.63\\
		pittsburg-bridges-REL-L&95.74$\pm$6.12 / 60.77$\pm$7.94 / 66.73$\pm$9.68&95.46$\pm$6.63 / 58.54$\pm$8.35 / 65.17$\pm$10.24\\
		pittsburg-bridges-SPAN&90.81$\pm$10.15 / 66.45$\pm$8.50 / 69.47$\pm$8.63&90.39$\pm$10.63 / 65.10$\pm$9.12 / 68.19$\pm$9.41\\
		pittsburg-bridges-T-OR-D&90.78$\pm$12.97 / 52.28$\pm$9.76 / 55.19$\pm$9.48&90.22$\pm$13.80 / 50.27$\pm$10.05 / 53.26$\pm$9.86\\
		pittsburg-bridges-TYPE&90.85$\pm$10.82 / 58.52$\pm$7.04 / 62.03$\pm$9.07&90.30$\pm$11.59 / 57.00$\pm$7.43 / 60.68$\pm$9.59\\
		planning&63.15$\pm$6.75 / 53.08$\pm$9.09 / 54.08$\pm$8.77&61.81$\pm$7.10 / 51.88$\pm$9.48 / 52.96$\pm$9.10\\
		plant-margin&84.57$\pm$3.13 / 83.21$\pm$1.23 / 85.54$\pm$1.03&84.59$\pm$3.18 / 83.17$\pm$1.23 / 85.51$\pm$1.04\\
		plant-shape&90.25$\pm$3.49 / 44.64$\pm$1.50 / 50.24$\pm$3.12&90.13$\pm$3.53 / 44.37$\pm$1.49 / 50.06$\pm$3.14\\
		plant-texture&71.67$\pm$6.69 / 85.40$\pm$1.58 / 86.46$\pm$1.34&70.78$\pm$7.31 / 85.37$\pm$1.59 / 86.43$\pm$1.35\\
		post-operative&98.14$\pm$2.41 / 54.34$\pm$6.33 / 57.77$\pm$6.02&98.09$\pm$2.49 / 52.19$\pm$6.93 / 55.88$\pm$6.45\\
		primary-tumor&94.12$\pm$8.14 / 57.80$\pm$3.45 / 64.41$\pm$5.09&94.01$\pm$8.32 / 57.09$\pm$3.58 / 63.93$\pm$5.22\\
		ringnorm&67.98$\pm$6.19 / 71.86$\pm$1.78 / 73.68$\pm$2.28&67.94$\pm$6.20 / 72.46$\pm$1.71 / 73.64$\pm$2.28\\
		seeds&84.51$\pm$12.34 / 71.13$\pm$6.79 / 76.48$\pm$4.69&84.18$\pm$12.65 / 70.58$\pm$6.95 / 76.06$\pm$4.74\\
		semeion&80.80$\pm$8.40 / 93.24$\pm$1.64 / 93.44$\pm$1.68&80.79$\pm$8.42 / 93.22$\pm$1.66 / 93.43$\pm$1.69\\
		soybean&81.95$\pm$9.46 / 89.02$\pm$1.77 / 89.55$\pm$2.01&81.74$\pm$9.50 / 88.97$\pm$1.78 / 89.50$\pm$2.02\\
		spambase&68.57$\pm$10.70 / 50.83$\pm$1.38 / 54.94$\pm$2.03&68.60$\pm$10.79 / 50.64$\pm$1.45 / 54.85$\pm$2.02\\
		spect&91.64$\pm$6.00 / 57.47$\pm$5.84 / 58.87$\pm$5.78&91.44$\pm$6.23 / 56.44$\pm$5.98 / 57.89$\pm$5.88\\
		spectf&71.67$\pm$10.87 / 36.18$\pm$1.69 / 38.21$\pm$2.11&70.90$\pm$10.89 / 35.07$\pm$1.25 / 37.57$\pm$2.05\\
		statlog-australian-credit&93.67$\pm$7.03 / 49.28$\pm$3.19 / 52.72$\pm$3.65&93.57$\pm$7.15 / 48.77$\pm$3.17 / 52.28$\pm$3.64\\
		statlog-german-credit&87.68$\pm$9.16 / 51.93$\pm$1.47 / 54.52$\pm$1.72&87.57$\pm$9.23 / 51.55$\pm$1.47 / 54.19$\pm$1.72\\
		statlog-heart&95.05$\pm$3.95 / 65.43$\pm$7.67 / 66.20$\pm$7.78&95.00$\pm$3.99 / 64.66$\pm$7.86 / 65.42$\pm$8.03\\
		statlog-image&91.28$\pm$6.42 / 73.10$\pm$1.73 / 76.46$\pm$2.90&91.24$\pm$6.45 / 73.10$\pm$1.75 / 76.38$\pm$2.91\\
		statlog-landsat&90.77$\pm$4.95 / 66.40$\pm$3.06 / 66.36$\pm$3.03&90.76$\pm$4.96 / 66.39$\pm$3.06 / 66.31$\pm$3.04\\
		statlog-shuttle&98.75$\pm$0.24 / 84.39$\pm$1.91 / 91.08$\pm$2.70&98.74$\pm$0.24 / 85.08$\pm$2.24 / 91.08$\pm$2.70\\
		statlog-vehicle&72.05$\pm$9.57 / 48.83$\pm$1.39 / 51.39$\pm$2.45&71.83$\pm$9.62 / 48.41$\pm$1.41 / 51.02$\pm$2.50\\
		steel-plates&84.43$\pm$7.57 / 56.04$\pm$3.65 / 57.69$\pm$3.13&84.33$\pm$7.63 / 55.85$\pm$3.68 / 57.50$\pm$3.15\\
		synthetic-control&87.88$\pm$4.31 / 77.81$\pm$5.41 / 77.72$\pm$5.63&87.78$\pm$4.34 / 77.61$\pm$5.54 / 77.48$\pm$5.74\\
		teaching&86.20$\pm$14.72 / 58.62$\pm$8.18 / 61.98$\pm$8.69&85.83$\pm$15.16 / 57.37$\pm$8.57 / 61.00$\pm$8.95\\
		thyroid&88.61$\pm$3.12 / 63.71$\pm$1.51 / 79.67$\pm$2.98&88.38$\pm$3.14 / 67.55$\pm$2.47 / 79.40$\pm$3.02\\
		tic-tac-toe&90.04$\pm$12.53 / 49.99$\pm$2.29 / 58.84$\pm$10.38&89.96$\pm$12.64 / 43.95$\pm$3.45 / 58.60$\pm$10.45\\
		titanic&92.10$\pm$9.43 / 62.55$\pm$3.56 / 62.78$\pm$3.43&92.07$\pm$9.46 / 62.48$\pm$3.58 / 62.71$\pm$3.45\\
		trains&100.00$\pm$0.00 / 70.83$\pm$12.50 / 81.94$\pm$1.39&100.00$\pm$0.00 / 60.42$\pm$18.75 / 77.78$\pm$1.39\\
		twonorm&72.65$\pm$1.13 / 77.18$\pm$1.09 / 77.02$\pm$1.30&72.62$\pm$1.13 / 77.34$\pm$1.07 / 76.99$\pm$1.30\\
		vertebral-column-2clases&67.89$\pm$7.26 / 45.66$\pm$2.19 / 46.14$\pm$2.13&67.30$\pm$7.52 / 44.74$\pm$2.09 / 45.15$\pm$2.03\\
		vertebral-column-3clases&62.91$\pm$4.82 / 49.22$\pm$4.02 / 49.66$\pm$3.95&62.11$\pm$5.10 / 48.45$\pm$4.06 / 48.94$\pm$3.99\\
		wall-following&75.23$\pm$7.20 / 52.40$\pm$1.30 / 56.62$\pm$2.61&75.18$\pm$7.21 / 52.14$\pm$1.35 / 56.55$\pm$2.62\\
		waveform&74.14$\pm$8.22 / 53.93$\pm$2.21 / 53.37$\pm$2.06&74.12$\pm$8.24 / 53.89$\pm$2.22 / 53.32$\pm$2.06\\
		waveform-noise&61.04$\pm$2.24 / 57.76$\pm$1.67 / 57.44$\pm$1.92&61.00$\pm$2.25 / 57.72$\pm$1.67 / 57.38$\pm$1.93\\
		wine&77.86$\pm$3.14 / 78.82$\pm$2.75 / 79.50$\pm$2.50&77.25$\pm$3.33 / 78.60$\pm$2.65 / 78.78$\pm$2.51\\
		wine-quality-red&69.71$\pm$5.04 / 48.11$\pm$3.99 / 49.34$\pm$3.78&69.51$\pm$5.06 / 47.87$\pm$4.03 / 49.12$\pm$3.82\\
		wine-quality-white&68.57$\pm$6.34 / 46.25$\pm$1.83 / 47.11$\pm$1.53&68.51$\pm$6.35 / 46.15$\pm$1.82 / 47.02$\pm$1.53\\
		yeast&85.99$\pm$10.60 / 58.22$\pm$2.61 / 60.09$\pm$4.06&85.88$\pm$10.66 / 58.05$\pm$2.63 / 59.93$\pm$4.10\\
		zoo&97.38$\pm$5.44 / 87.52$\pm$7.03 / 88.92$\pm$7.45&97.26$\pm$5.73 / 86.85$\pm$7.88 / 88.36$\pm$8.20\\
		\bottomrule
\end{longtable}}

{\scriptsize
	\begin{longtable}{l c c}
		\caption{Comparison between RED and Counterparts on adversarial sample detection (mean$\pm$std over 10 runs)} \label{app:table_UCI125_adversarial}\\
		\toprule
		\multirow{2}{*}{Dataset} & AP-adversarial(\%) & AUPR-adversarial(\%)\\
		{}	& RED-variance / MCP baseline / RED-mean & RED-variance / MCP baseline / RED-mean\\
		\midrule
		\endfirsthead
		\toprule
		\multirow{2}{*}{Dataset} & AP-adversarial(\%) & AUPR-adversarial(\%)\\
		{}	& RED-variance / MCP baseline / RED-mean & RED-variance / MCP baseline / RED-mean\\
		\midrule
		\endhead
		\\\emph{Continued on next page.} & & \\
		\endfoot
		\\\emph{Continued from previous page.} & & \\
		\endlastfoot
		abalone&82.29$\pm$16.60 / 50.00$\pm$0.00 / 31.25$\pm$0.69&81.88$\pm$16.98 / 25.00$\pm$0.00 / 30.98$\pm$0.34\\
		acute-inflammation&87.52$\pm$5.08 / 50.00$\pm$0.00 / 36.64$\pm$1.87&87.30$\pm$5.16 / 25.00$\pm$0.00 / 35.21$\pm$1.72\\
		acute-nephritis&91.47$\pm$3.88 / 50.00$\pm$0.00 / 38.04$\pm$4.42&91.31$\pm$3.97 / 25.00$\pm$0.00 / 36.56$\pm$4.42\\
		adult&79.20$\pm$9.01 / 50.00$\pm$0.00 / 37.79$\pm$2.58&79.14$\pm$8.98 / 25.00$\pm$0.00 / 37.71$\pm$2.61\\
		annealing&74.83$\pm$5.64 / 50.00$\pm$0.00 / 38.85$\pm$2.84&73.69$\pm$5.82 / 25.00$\pm$0.00 / 37.39$\pm$3.24\\
		arrhythmia&71.19$\pm$9.17 / 50.00$\pm$0.00 / 34.37$\pm$1.64&69.56$\pm$9.60 / 25.00$\pm$0.00 / 32.38$\pm$1.74\\
		audiology-std&68.10$\pm$6.52 / 50.00$\pm$0.00 / 34.33$\pm$1.29&67.10$\pm$6.84 / 25.00$\pm$0.00 / 33.21$\pm$1.37\\
		balance-scale&76.13$\pm$4.77 / 50.00$\pm$0.00 / 66.60$\pm$8.53&75.76$\pm$4.84 / 25.00$\pm$0.00 / 66.17$\pm$8.54\\
		balloons&90.37$\pm$10.15 / 50.00$\pm$0.00 / 61.60$\pm$19.27&88.95$\pm$11.63 / 25.00$\pm$0.00 / 52.92$\pm$23.70\\
		bank&96.80$\pm$1.03 / 50.00$\pm$0.00 / 40.97$\pm$11.31&96.80$\pm$1.04 / 25.00$\pm$0.00 / 40.92$\pm$11.31\\
		bioconcentration&95.04$\pm$4.88 / 50.00$\pm$0.00 / 42.40$\pm$5.27&94.78$\pm$5.32 / 25.00$\pm$0.00 / 40.10$\pm$4.28\\
		blood&75.93$\pm$12.33 / 50.00$\pm$0.00 / 30.99$\pm$0.05&75.46$\pm$12.58 / 25.00$\pm$0.00 / 30.68$\pm$0.00\\
		breast-cancer&88.40$\pm$12.51 / 50.00$\pm$0.00 / 32.05$\pm$0.85&88.09$\pm$12.83 / 25.00$\pm$0.00 / 31.02$\pm$0.24\\
		breast-cancer-wisc&61.42$\pm$3.98 / 50.00$\pm$0.00 / 41.86$\pm$6.71&60.81$\pm$3.98 / 25.00$\pm$0.00 / 41.33$\pm$6.64\\
		breast-cancer-wisc-diag&58.41$\pm$2.24 / 50.00$\pm$0.00 / 47.50$\pm$7.89&57.97$\pm$2.40 / 25.00$\pm$0.00 / 47.04$\pm$7.83\\
		breast-cancer-wisc-prog&93.18$\pm$7.24 / 50.00$\pm$0.00 / 34.63$\pm$1.71&92.99$\pm$7.56 / 25.00$\pm$0.00 / 32.85$\pm$1.03\\
		breast-tissue&70.92$\pm$8.00 / 50.00$\pm$0.00 / 33.41$\pm$0.97&69.28$\pm$8.21 / 25.00$\pm$0.00 / 31.82$\pm$0.64\\
		car&78.72$\pm$5.27 / 50.00$\pm$0.00 / 54.68$\pm$9.10&78.50$\pm$5.25 / 25.00$\pm$0.00 / 54.15$\pm$9.60\\
		cardiotocography-10clases&68.63$\pm$4.28 / 50.00$\pm$0.00 / 34.28$\pm$0.94&68.53$\pm$4.30 / 25.00$\pm$0.00 / 34.20$\pm$0.93\\
		cardiotocography-3clases&87.37$\pm$6.47 / 50.00$\pm$0.00 / 51.54$\pm$7.78&87.00$\pm$6.65 / 25.00$\pm$0.00 / 51.28$\pm$7.95\\
		chess-krvk&75.63$\pm$5.23 / 50.00$\pm$0.00 / 31.90$\pm$0.24&75.62$\pm$5.23 / 25.00$\pm$0.00 / 31.88$\pm$0.25\\
		chess-krvkp&60.18$\pm$4.97 / 50.00$\pm$0.00 / 51.07$\pm$6.63&56.88$\pm$2.79 / 25.00$\pm$0.00 / 48.88$\pm$5.20\\
		climate&84.22$\pm$9.18 / 50.00$\pm$0.00 / 38.39$\pm$2.79&83.51$\pm$9.96 / 25.00$\pm$0.00 / 36.69$\pm$1.56\\
		congressional-voting&72.04$\pm$14.03 / 50.00$\pm$0.00 / 35.03$\pm$0.77&69.59$\pm$15.23 / 25.00$\pm$0.00 / 30.49$\pm$0.10\\
		conn-bench-sonar-mines-rocks&58.24$\pm$9.74 / 50.00$\pm$0.00 / 33.42$\pm$2.91&57.17$\pm$9.78 / 25.00$\pm$0.00 / 32.63$\pm$2.91\\
		conn-bench-vowel-deterding&57.23$\pm$4.07 / 50.00$\pm$0.00 / 39.97$\pm$3.24&56.82$\pm$4.16 / 25.00$\pm$0.00 / 39.74$\pm$3.21\\
		connect-4&85.56$\pm$8.53 / 50.00$\pm$0.00 / 38.84$\pm$2.17&85.39$\pm$8.81 / 25.00$\pm$0.00 / 38.01$\pm$1.77\\
		contrac&89.59$\pm$16.29 / 50.00$\pm$0.00 / 31.34$\pm$0.72&89.53$\pm$16.37 / 25.00$\pm$0.00 / 30.82$\pm$0.09\\
		credit-approval&68.12$\pm$10.10 / 50.00$\pm$0.00 / 33.69$\pm$0.92&67.47$\pm$10.05 / 25.00$\pm$0.00 / 32.63$\pm$1.07\\
		cylinder-bands&70.11$\pm$13.95 / 50.00$\pm$0.00 / 35.28$\pm$2.27&68.85$\pm$15.04 / 25.00$\pm$0.00 / 33.76$\pm$1.92\\
		dermatology&65.38$\pm$4.23 / 50.00$\pm$0.00 / 42.58$\pm$6.20&63.96$\pm$4.13 / 25.00$\pm$0.00 / 41.61$\pm$6.41\\
		echocardiogram&66.62$\pm$12.36 / 50.00$\pm$0.00 / 34.48$\pm$2.00&65.20$\pm$12.92 / 25.00$\pm$0.00 / 32.91$\pm$1.40\\
		ecoli&69.15$\pm$7.45 / 50.00$\pm$0.00 / 33.64$\pm$2.24&68.45$\pm$7.56 / 25.00$\pm$0.00 / 33.16$\pm$2.15\\
		energy-y1&74.51$\pm$8.01 / 50.00$\pm$0.00 / 46.99$\pm$6.10&74.25$\pm$8.06 / 25.00$\pm$0.00 / 46.61$\pm$6.15\\
		energy-y2&70.30$\pm$10.32 / 50.00$\pm$0.00 / 43.50$\pm$5.93&69.98$\pm$10.30 / 25.00$\pm$0.00 / 43.22$\pm$5.91\\
		fertility&98.25$\pm$1.45 / 50.00$\pm$0.00 / 41.95$\pm$10.58&98.20$\pm$1.49 / 25.00$\pm$0.00 / 38.26$\pm$9.34\\
		flags&82.17$\pm$10.49 / 50.00$\pm$0.00 / 33.18$\pm$1.28&81.61$\pm$10.80 / 25.00$\pm$0.00 / 31.28$\pm$0.56\\
		glass&77.23$\pm$8.67 / 50.00$\pm$0.00 / 33.24$\pm$0.86&76.39$\pm$9.18 / 25.00$\pm$0.00 / 31.86$\pm$0.40\\
		haberman-survival&75.17$\pm$13.90 / 50.00$\pm$0.00 / 31.32$\pm$0.39&74.51$\pm$14.18 / 25.00$\pm$0.00 / 30.77$\pm$0.20\\
		hayes-roth&74.57$\pm$16.18 / 50.00$\pm$0.00 / 33.61$\pm$1.93&73.49$\pm$16.94 / 25.00$\pm$0.00 / 31.94$\pm$1.55\\
		heart-cleveland&88.01$\pm$13.13 / 50.00$\pm$0.00 / 33.36$\pm$1.65&87.81$\pm$13.33 / 25.00$\pm$0.00 / 31.74$\pm$0.56\\
		heart-hungarian&68.16$\pm$10.33 / 50.00$\pm$0.00 / 34.66$\pm$3.34&66.96$\pm$10.91 / 25.00$\pm$0.00 / 33.02$\pm$2.99\\
		heart-switzerland&96.64$\pm$3.48 / 50.00$\pm$0.00 / 36.16$\pm$1.80&96.12$\pm$4.16 / 25.00$\pm$0.00 / 30.91$\pm$0.28\\
		heart-va&93.95$\pm$4.60 / 50.00$\pm$0.00 / 33.91$\pm$1.79&93.47$\pm$5.05 / 25.00$\pm$0.00 / 30.82$\pm$0.13\\
		hepatitis&91.75$\pm$6.67 / 50.00$\pm$0.00 / 39.52$\pm$3.68&91.51$\pm$6.88 / 25.00$\pm$0.00 / 37.31$\pm$4.38\\
		hill-valley&82.36$\pm$18.18 / 50.00$\pm$0.00 / 31.62$\pm$0.91&82.28$\pm$18.26 / 25.00$\pm$0.00 / 31.14$\pm$0.50\\
		horse-colic&69.38$\pm$13.84 / 50.00$\pm$0.00 / 37.48$\pm$3.81&68.39$\pm$14.69 / 25.00$\pm$0.00 / 36.16$\pm$3.85\\
		ilpd-indian-liver&81.30$\pm$16.46 / 50.00$\pm$0.00 / 31.89$\pm$0.40&81.18$\pm$16.58 / 25.00$\pm$0.00 / 31.60$\pm$0.43\\
		image-segmentation&68.08$\pm$7.14 / 50.00$\pm$0.00 / 46.76$\pm$2.06&67.96$\pm$7.13 / 25.00$\pm$0.00 / 46.58$\pm$1.96\\
		ionosphere&62.59$\pm$8.10 / 50.00$\pm$0.00 / 40.37$\pm$5.35&61.98$\pm$8.13 / 25.00$\pm$0.00 / 39.55$\pm$5.20\\
		iris&66.03$\pm$13.61 / 50.00$\pm$0.00 / 44.15$\pm$10.31&64.86$\pm$14.01 / 25.00$\pm$0.00 / 42.90$\pm$10.32\\
		led-display&64.69$\pm$6.48 / 50.00$\pm$0.00 / 31.75$\pm$0.54&63.77$\pm$6.59 / 25.00$\pm$0.00 / 30.78$\pm$0.22\\
		lenses&90.15$\pm$13.05 / 50.00$\pm$0.00 / 42.76$\pm$5.29&88.90$\pm$14.81 / 25.00$\pm$0.00 / 33.57$\pm$3.91\\
		letter&64.40$\pm$3.32 / 50.00$\pm$0.00 / 40.53$\pm$0.98&64.38$\pm$3.32 / 25.00$\pm$0.00 / 40.50$\pm$0.97\\
		libras&58.83$\pm$3.65 / 50.00$\pm$0.00 / 33.24$\pm$0.72&58.11$\pm$3.71 / 25.00$\pm$0.00 / 32.77$\pm$0.70\\
		low-res-spect&81.18$\pm$5.65 / 50.00$\pm$0.00 / 45.85$\pm$8.44&80.60$\pm$5.91 / 25.00$\pm$0.00 / 44.81$\pm$8.56\\
		lung-cancer&86.08$\pm$17.22 / 50.00$\pm$0.00 / 40.30$\pm$2.63&83.82$\pm$20.46 / 25.00$\pm$0.00 / 31.73$\pm$2.31\\
		lymphography&84.63$\pm$11.97 / 50.00$\pm$0.00 / 40.57$\pm$3.52&83.29$\pm$13.04 / 25.00$\pm$0.00 / 38.41$\pm$3.82\\
		magic&57.43$\pm$6.63 / 50.00$\pm$0.00 / 32.90$\pm$0.73&57.38$\pm$6.61 / 25.00$\pm$0.00 / 32.80$\pm$0.83\\
		mammographic&65.57$\pm$14.20 / 50.00$\pm$0.00 / 31.05$\pm$0.34&65.30$\pm$14.31 / 25.00$\pm$0.00 / 30.89$\pm$0.34\\
		messidor&81.71$\pm$9.63 / 50.00$\pm$0.00 / 38.45$\pm$1.99&81.18$\pm$10.32 / 25.00$\pm$0.00 / 37.43$\pm$2.06\\
		miniboone&74.95$\pm$4.30 / 50.00$\pm$0.00 / 44.62$\pm$3.75&73.73$\pm$3.33 / 25.00$\pm$0.00 / 43.57$\pm$3.17\\
		molec-biol-promoter&63.69$\pm$15.17 / 50.00$\pm$0.00 / 34.43$\pm$2.56&62.22$\pm$15.78 / 25.00$\pm$0.00 / 32.34$\pm$2.68\\
		molec-biol-splice&77.51$\pm$10.87 / 50.00$\pm$0.00 / 38.85$\pm$4.16&76.17$\pm$12.14 / 25.00$\pm$0.00 / 36.63$\pm$3.57\\
		monks-1&55.16$\pm$2.17 / 50.00$\pm$0.00 / 50.97$\pm$7.31&53.93$\pm$3.21 / 25.00$\pm$0.00 / 49.72$\pm$6.43\\
		monks-2&58.36$\pm$13.80 / 50.00$\pm$0.00 / 34.04$\pm$1.85&57.99$\pm$13.93 / 25.00$\pm$0.00 / 33.78$\pm$1.84\\
		monks-3&57.76$\pm$4.07 / 50.00$\pm$0.00 / 42.44$\pm$3.77&57.26$\pm$4.13 / 25.00$\pm$0.00 / 41.98$\pm$3.73\\
		mushroom&67.93$\pm$1.65 / 50.00$\pm$0.00 / 59.28$\pm$2.56&67.90$\pm$1.66 / 45.00$\pm$24.49 / 59.24$\pm$2.56\\
		musk-1&62.71$\pm$2.54 / 50.00$\pm$0.00 / 40.57$\pm$7.14&61.68$\pm$2.64 / 25.00$\pm$0.00 / 38.57$\pm$6.99\\
		musk-2&90.10$\pm$2.91 / 50.00$\pm$0.00 / 66.96$\pm$17.08&91.35$\pm$4.69 / 25.00$\pm$0.00 / 59.27$\pm$19.68\\
		nursery&82.40$\pm$4.02 / 50.00$\pm$0.00 / 60.87$\pm$12.66&82.39$\pm$4.02 / 25.00$\pm$0.00 / 60.85$\pm$12.66\\
		oocytes\_merluccius\_nucleus\_4d&73.72$\pm$13.34 / 50.00$\pm$0.00 / 34.69$\pm$0.98&73.33$\pm$13.67 / 25.00$\pm$0.00 / 33.72$\pm$1.63\\
		oocytes\_merluccius\_states\_2f&76.71$\pm$6.36 / 50.00$\pm$0.00 / 46.03$\pm$4.93&76.22$\pm$6.72 / 25.00$\pm$0.00 / 45.66$\pm$4.78\\
		oocytes\_trisopterus\_nucleus\_2f&67.67$\pm$15.70 / 50.00$\pm$0.00 / 34.20$\pm$1.97&66.70$\pm$15.97 / 25.00$\pm$0.00 / 32.81$\pm$1.28\\
		oocytes\_trisopterus\_states\_5b&74.64$\pm$9.22 / 50.00$\pm$0.00 / 38.82$\pm$3.28&74.13$\pm$9.66 / 25.00$\pm$0.00 / 38.19$\pm$3.53\\
		optical&55.66$\pm$1.99 / 50.00$\pm$0.00 / 41.89$\pm$1.65&54.98$\pm$1.78 / 25.00$\pm$0.00 / 41.79$\pm$1.65\\
		ozone&99.40$\pm$0.14 / 50.00$\pm$0.00 / 57.84$\pm$19.77&99.40$\pm$0.13 / 25.00$\pm$0.00 / 53.85$\pm$20.73\\
		page-blocks&85.76$\pm$2.72 / 50.00$\pm$0.00 / 62.95$\pm$9.31&85.66$\pm$2.71 / 25.00$\pm$0.00 / 62.88$\pm$9.33\\
		parkinsons&68.29$\pm$12.63 / 50.00$\pm$0.00 / 40.64$\pm$3.51&67.52$\pm$13.06 / 25.00$\pm$0.00 / 39.51$\pm$3.33\\
		pendigits&65.20$\pm$10.65 / 50.00$\pm$0.00 / 46.03$\pm$3.60&65.68$\pm$10.76 / 25.00$\pm$0.00 / 45.81$\pm$3.59\\
		phishing&93.81$\pm$4.90 / 50.00$\pm$0.00 / 40.41$\pm$2.56&93.58$\pm$5.31 / 25.00$\pm$0.00 / 40.00$\pm$2.59\\
		pima&71.85$\pm$18.93 / 50.00$\pm$0.00 / 31.70$\pm$0.59&71.62$\pm$19.07 / 25.00$\pm$0.00 / 31.24$\pm$0.56\\
		pittsburg-bridges-MATERIAL&92.85$\pm$7.91 / 50.00$\pm$0.00 / 41.62$\pm$5.77&92.60$\pm$8.13 / 25.00$\pm$0.00 / 38.30$\pm$5.53\\
		pittsburg-bridges-REL-L&88.67$\pm$13.51 / 50.00$\pm$0.00 / 33.51$\pm$1.20&88.20$\pm$14.11 / 25.00$\pm$0.00 / 31.44$\pm$0.93\\
		pittsburg-bridges-SPAN&88.45$\pm$11.20 / 50.00$\pm$0.00 / 34.72$\pm$2.90&87.52$\pm$11.62 / 25.00$\pm$0.00 / 30.98$\pm$0.34\\
		pittsburg-bridges-T-OR-D&89.89$\pm$14.29 / 50.00$\pm$0.00 / 36.92$\pm$4.57&89.49$\pm$14.94 / 25.00$\pm$0.00 / 34.61$\pm$4.73\\
		pittsburg-bridges-TYPE&75.32$\pm$14.52 / 50.00$\pm$0.00 / 33.57$\pm$1.46&73.58$\pm$14.90 / 25.00$\pm$0.00 / 31.43$\pm$0.50\\
		planning&93.22$\pm$10.61 / 50.00$\pm$0.00 / 31.47$\pm$0.15&93.08$\pm$10.88 / 25.00$\pm$0.00 / 30.74$\pm$0.13\\
		plant-margin&55.85$\pm$1.56 / 50.00$\pm$0.00 / 32.93$\pm$0.38&56.25$\pm$1.68 / 25.00$\pm$0.00 / 32.75$\pm$0.37\\
		plant-shape&58.21$\pm$2.10 / 50.00$\pm$0.00 / 31.52$\pm$0.24&58.02$\pm$2.14 / 25.00$\pm$0.00 / 31.43$\pm$0.24\\
		plant-texture&54.14$\pm$1.85 / 50.00$\pm$0.00 / 32.35$\pm$0.43&54.24$\pm$1.68 / 25.00$\pm$0.00 / 32.24$\pm$0.43\\
		post-operative&90.65$\pm$8.00 / 50.00$\pm$0.00 / 34.10$\pm$2.58&90.18$\pm$8.32 / 25.00$\pm$0.00 / 30.77$\pm$0.28\\
		primary-tumor&76.22$\pm$9.80 / 50.00$\pm$0.00 / 32.30$\pm$0.76&75.15$\pm$10.20 / 25.00$\pm$0.00 / 31.08$\pm$0.27\\
		ringnorm&58.38$\pm$2.62 / 50.00$\pm$0.00 / 45.19$\pm$4.75&57.46$\pm$2.38 / 25.00$\pm$0.00 / 44.37$\pm$3.93\\
		seeds&64.50$\pm$10.51 / 50.00$\pm$0.00 / 43.79$\pm$8.08&63.25$\pm$10.54 / 25.00$\pm$0.00 / 42.80$\pm$7.92\\
		semeion&59.44$\pm$2.23 / 50.00$\pm$0.00 / 39.29$\pm$1.88&57.00$\pm$2.75 / 25.00$\pm$0.00 / 36.98$\pm$1.67\\
		soybean&59.34$\pm$5.89 / 50.00$\pm$0.00 / 34.96$\pm$1.73&58.97$\pm$5.96 / 25.00$\pm$0.00 / 34.61$\pm$1.78\\
		spambase&57.38$\pm$4.15 / 50.00$\pm$0.00 / 38.44$\pm$2.41&55.89$\pm$3.93 / 25.00$\pm$0.00 / 37.21$\pm$2.41\\
		spect&96.57$\pm$2.08 / 50.00$\pm$0.00 / 34.48$\pm$2.94&96.40$\pm$2.11 / 25.00$\pm$0.00 / 31.32$\pm$0.75\\
		spectf&87.97$\pm$9.00 / 50.00$\pm$0.00 / 33.95$\pm$1.79&87.80$\pm$9.15 / 25.00$\pm$0.00 / 33.26$\pm$1.80\\
		statlog-australian-credit&95.53$\pm$8.34 / 50.00$\pm$0.00 / 31.97$\pm$2.07&95.50$\pm$8.38 / 25.00$\pm$0.00 / 30.77$\pm$0.08\\
		statlog-german-credit&91.57$\pm$2.51 / 50.00$\pm$0.00 / 32.95$\pm$2.42&91.42$\pm$2.51 / 25.00$\pm$0.00 / 31.32$\pm$0.35\\
		statlog-heart&67.93$\pm$14.05 / 50.00$\pm$0.00 / 35.23$\pm$2.39&66.88$\pm$13.96 / 25.00$\pm$0.00 / 34.08$\pm$1.89\\
		statlog-image&66.45$\pm$8.06 / 50.00$\pm$0.00 / 44.83$\pm$2.63&66.33$\pm$8.06 / 25.00$\pm$0.00 / 44.72$\pm$2.63\\
		statlog-landsat&68.63$\pm$4.95 / 50.00$\pm$0.00 / 39.26$\pm$1.90&68.58$\pm$4.95 / 25.00$\pm$0.00 / 39.11$\pm$1.95\\
		statlog-shuttle&99.35$\pm$0.19 / 50.00$\pm$0.00 / 71.57$\pm$20.88&99.32$\pm$0.22 / 25.00$\pm$0.00 / 71.49$\pm$20.83\\
		statlog-vehicle&63.62$\pm$7.78 / 50.00$\pm$0.00 / 35.04$\pm$1.41&63.30$\pm$7.75 / 25.00$\pm$0.00 / 34.71$\pm$1.50\\
		steel-plates&63.97$\pm$8.77 / 50.00$\pm$0.00 / 33.30$\pm$1.04&63.81$\pm$8.78 / 25.00$\pm$0.00 / 33.06$\pm$1.09\\
		synthetic-control&55.00$\pm$1.43 / 50.00$\pm$0.00 / 40.00$\pm$2.98&54.18$\pm$1.25 / 25.00$\pm$0.00 / 39.44$\pm$3.11\\
		teaching&83.62$\pm$17.31 / 50.00$\pm$0.00 / 32.65$\pm$1.42&82.88$\pm$17.90 / 25.00$\pm$0.00 / 30.75$\pm$0.12\\
		thyroid&90.25$\pm$2.52 / 50.00$\pm$0.00 / 82.43$\pm$2.70&90.17$\pm$2.55 / 25.00$\pm$0.00 / 82.34$\pm$2.79\\
		tic-tac-toe&64.84$\pm$6.06 / 50.00$\pm$0.00 / 49.16$\pm$10.24&63.61$\pm$5.92 / 25.00$\pm$0.00 / 48.45$\pm$10.18\\
		titanic&81.73$\pm$12.73 / 50.00$\pm$0.00 / 34.62$\pm$0.85&78.65$\pm$14.12 / 25.00$\pm$0.00 / 30.64$\pm$0.30\\
		trains&68.06$\pm$15.28 / 50.00$\pm$0.00 / 41.67$\pm$0.00&61.11$\pm$18.06 / 25.00$\pm$0.00 / 29.17$\pm$0.00\\
		twonorm&62.77$\pm$4.51 / 50.00$\pm$0.00 / 43.06$\pm$1.54&62.11$\pm$4.11 / 25.00$\pm$0.00 / 42.81$\pm$1.77\\
		vertebral-column-2clases&68.33$\pm$14.00 / 50.00$\pm$0.00 / 34.72$\pm$1.75&67.77$\pm$14.32 / 25.00$\pm$0.00 / 34.21$\pm$1.72\\
		vertebral-column-3clases&64.11$\pm$8.66 / 50.00$\pm$0.00 / 34.50$\pm$1.13&63.45$\pm$8.81 / 25.00$\pm$0.00 / 33.80$\pm$1.37\\
		wall-following&63.60$\pm$4.25 / 50.00$\pm$0.00 / 40.26$\pm$3.14&63.53$\pm$4.23 / 25.00$\pm$0.00 / 40.22$\pm$3.14\\
		waveform&55.60$\pm$8.02 / 50.00$\pm$0.00 / 36.50$\pm$1.06&55.42$\pm$8.05 / 25.00$\pm$0.00 / 36.30$\pm$0.97\\
		waveform-noise&56.22$\pm$9.33 / 50.00$\pm$0.00 / 36.78$\pm$2.18&54.53$\pm$9.90 / 25.00$\pm$0.00 / 34.02$\pm$0.49\\
		wine&66.01$\pm$5.38 / 50.00$\pm$0.00 / 52.04$\pm$12.94&64.74$\pm$5.36 / 25.00$\pm$0.00 / 50.76$\pm$12.72\\
		wine-quality-red&91.62$\pm$10.31 / 50.00$\pm$0.00 / 31.74$\pm$1.73&91.47$\pm$10.55 / 25.00$\pm$0.00 / 30.82$\pm$0.09\\
		wine-quality-white&94.70$\pm$6.31 / 50.00$\pm$0.00 / 31.17$\pm$0.70&94.47$\pm$6.75 / 25.00$\pm$0.00 / 30.70$\pm$0.01\\
		yeast&77.94$\pm$7.43 / 50.00$\pm$0.00 / 31.42$\pm$0.90&77.60$\pm$7.64 / 25.00$\pm$0.00 / 30.78$\pm$0.12\\
		zoo&76.04$\pm$7.49 / 50.00$\pm$0.00 / 41.41$\pm$6.60&73.72$\pm$8.18 / 25.00$\pm$0.00 / 38.18$\pm$6.51\\
		\bottomrule
\end{longtable}}

\end{document}